\pdfoutput=1

\documentclass[11pt]{article}

\usepackage[]{EMNLP2022}

\usepackage{times}
\usepackage{latexsym}

\usepackage[T1]{fontenc}

\usepackage[utf8]{inputenc}

\usepackage{microtype}

\usepackage{graphicx}
\usepackage{booktabs} 
\usepackage{caption}
\usepackage{subcaption}

\usepackage{hyperref}

\usepackage[utf8]{inputenc}
\usepackage{amsmath,amsthm,amssymb,mdframed}
\usepackage{multirow}
\usepackage{xspace}
\usepackage{etoolbox}
\usepackage{stfloats}
\usepackage{latexsym}

\usepackage{tabularx}

\usepackage{graphicx}
\usepackage{enumitem}
\usepackage{floatrow}
\usepackage{bm}
\usepackage{bbm}

\usepackage{blindtext}

\usepackage{wrapfig}
\usepackage{booktabs}
\usepackage{placeins}
\usepackage{xspace}
\usepackage{tabularx}

\usepackage[T1]{fontenc}

\usepackage[utf8]{inputenc}

\usepackage{amsmath}
\usepackage{amssymb}
\usepackage{mathtools}
\usepackage{amsthm}
\usepackage{amsfonts}
\usepackage[capitalize,noabbrev]{cleveref}
\usepackage{caption}

\theoremstyle{plain}
\newtheorem{theorem}{Theorem}[section]
\newtheorem{proposition}[theorem]{Proposition}

\theoremstyle{definition}
\newtheorem{definition}[theorem]{Definition}

\theoremstyle{remark}

\usepackage[textsize=tiny]{todonotes}

\newcommand{\system}[1]{\textsc{#1}\xspace}
\newcommand{\data}[1]{\textsc{#1}\xspace}

\newcommand{\tacred}{\data{TACRED}\xspace}
\newcommand{\empathe}{\data{Empathetic}\xspace}
\newcommand{\persona}{\data{Personality}\xspace}
\newcommand{\fewrel}{\data{FewRel}\xspace}

\newcommand{\noaug}{\system{No Aug}\xspace}
\newcommand{\eda}{\system{EDA}\xspace}
\newcommand{\cbert}{\system{BERT}\xspace}
\newcommand{\back}{\system{RTT}\xspace}
\newcommand{\lamb}{\system{LAMBDA}\xspace}
\newcommand{\ex}{\system{Ex2}\xspace}

\newcommand{\casuallens}{\system{Casual-Lens}\xspace}

\newcommand{\optimus}{\system{Optimus}\xspace}
\newcommand{\rvae}{\system{R-VAE-AVG}\xspace}
\newcommand{\cf}{\system{R-VAE-CF}\xspace}
\newcommand{\zf}{\system{ZF}\xspace}

\newcommand{\adv}{\system{gan}\xspace}
\newcommand{\mmd}{\system{mmd}\xspace}
\newcommand{\hsic}{\system{hsic}\xspace}
\newcommand{\idel}{\system{idel}\xspace}

\newcommand{\auto}{\system{vae-dprior (ae)}}
\newcommand{\ivae}{\system{vae-dprior}\xspace}
\newcommand{\vae}{\system{vae (uncond)}}
\newcommand{\ivaelgvar}{\system{vae-dprior (lgvar)}}
\newcommand{\vqvae}{\system{vq-vae}\xspace}
\newcommand{\cvae}{\system{c-vae}\xspace}
\newcommand{\vampvae}{\system{vamp-vae}\xspace}
\newcommand{\ivaerand}{\system{vae-dprior (rand)}}
\newcommand{\ivaerandft}{\system{vae-dprior (rand-ft)}}
\newcommand{\ivaegauss}{\system{vae-dprior (gauss)}}

\newcommand{\lencoder}{\system{bert + label encoder}\xspace}
\newcommand{\perplexity}{\system{perplexity}\xspace}
\newcommand{\random}{\system{random}\xspace}
\newcommand{\bert}{\system{bert}\xspace}

\newcommand{\ourmodel}{\system{VAE-DPrior}}
\newcommand{\encoder}{\system{BERT}}
\newcommand{\decoder}{\system{GPT2}}

\newcommand\independent{\protect\mathpalette{\protect\independenT}{\perp}}
\def\independenT#1#2{\mathrel{\rlap{$#1#2$}\mkern2mu{#1#2}}}

\usepackage{amsthm,amsmath,amsfonts,bm,xspace}
\usepackage{color}

\def\eqref#1{(\ref{#1})}

\def\1{\bm{1}}
\newcommand{\train}{\mathcal{D}}

\def\rmH{{\mathbf{H}}}
\def\rmI{{\mathbf{I}}}

\def\rmK{{\mathbf{K}}}

\def\rmV{{\mathbf{V}}}
\def\rmW{{\mathbf{W}}}
\def\rmX{{\mathbf{X}}}

\def\rmZ{{\mathbf{Z}}}

\def\vtheta{{\bm{\theta}}}

\def\vc{{\bm{c}}}

\def\vh{{\bm{h}}}

\def\vv{{\bm{v}}}

\def\vx{{\bm{x}}}
\def\vy{{\bm{y}}}
\def\vz{{\bm{z}}}

\DeclareMathAlphabet{\mathsfit}{\encodingdefault}{\sfdefault}{m}{sl}
\SetMathAlphabet{\mathsfit}{bold}{\encodingdefault}{\sfdefault}{bx}{n}

\def\gD{{\mathcal{D}}}

\def\gL{{\mathcal{L}}}

\def\gY{{\mathcal{Y}}}
\def\gZ{{\mathcal{Z}}}

\newcommand{\KL}{\mathbb{D}_{\mathrm{KL}}}
\newcommand{\diver}{\mathbb{D}}

\newcommand{\vAlpha}{\bm{\alpha}}
\newcommand{\vBeta}{\bm{\beta}}
\newcommand{\vTheta}{\bm{\theta}}
\newcommand{\vPhi}{\bm{\phi}}
\newcommand{\pT}{p_{\vTheta}}
\newcommand{\pTc}{p_{\vTheta_c}}
\newcommand{\pTy}{p_{\vTheta_y}}
\newcommand{\qP}{q_{\vPhi}}
\newcommand{\cons}{\vAlpha_c, \vBeta_y}
\newcommand{\vMu}{\bm{\mu}}
\newcommand{\vSigma}{\bm{\sigma}}

\DeclareMathOperator*{\argmax}{arg\,max}

\title{Variational Autoencoder with Disentanglement Priors for Low-Resource Task-Specific Natural
Language Generation}

\newcommand{\monash}{\triangle}
\newcommand{\mel}{\heartsuit}
\newcommand{\byd}{\diamondsuit}

\author{Zhuang Li$^{\monash}$, Lizhen Qu\Thanks{~~corresponding author}$\ \ ^{,\monash}$, Qiongkai Xu$^{\mel}$ \\ {\bf Tongtong Wu$^{\monash}$, Tianyang Zhan\thanks{~~Most of this author's work was finished when he was with Monash University.}$\ \ ^{,\byd}$,  Gholamreza Haffari$^{\monash}$} \\
  Monash University, Australia$^{\monash,\byd}$, 
  The University of Melbourne, Australia$^{\mel}$\\
  \texttt{firstname.lastname@monash.edu}$^{\monash}$,
  \texttt{firstname.lastname@unimelb.edu.au}$^{\mel}$, \\
  \texttt{tzha225@student.monash.edu}$^{\byd}$ \\
  }

\begin{document}
 \abovedisplayskip=0.04pt
\abovedisplayshortskip=0.04pt
\belowdisplayskip=0.04pt
\belowdisplayshortskip=0.04pt
\maketitle
\begin{abstract}
In this paper, we propose a variational autoencoder with disentanglement priors, \ourmodel, for task-specific natural language generation with none or a handful of task-specific labeled examples. In order to tackle compositional generalization across tasks, our model performs disentangled representation learning by introducing a conditional prior for the latent content space and another conditional prior for the latent label space. Both types of priors satisfy a novel property called $\epsilon$-disentangled. We show both empirically and theoretically that the novel priors can disentangle representations even without specific regularizations as in the prior work. The content prior enables directly sampling diverse content representations from the content space learned from the seen tasks, and fuse them with the representations of novel tasks for generating semantically diverse texts in the low-resource settings. Our extensive experiments demonstrate the superior performance of our model over competitive baselines in terms of i) data augmentation in continuous zero/few-shot learning, and ii) text style transfer in the few-shot setting. The code is available at \url{https://github.com/zhuang-li/VAE-DPrior}.
\end{abstract}

\section{Introduction}

\begin{figure}[t]
    \centering
    \includegraphics[width=\linewidth]{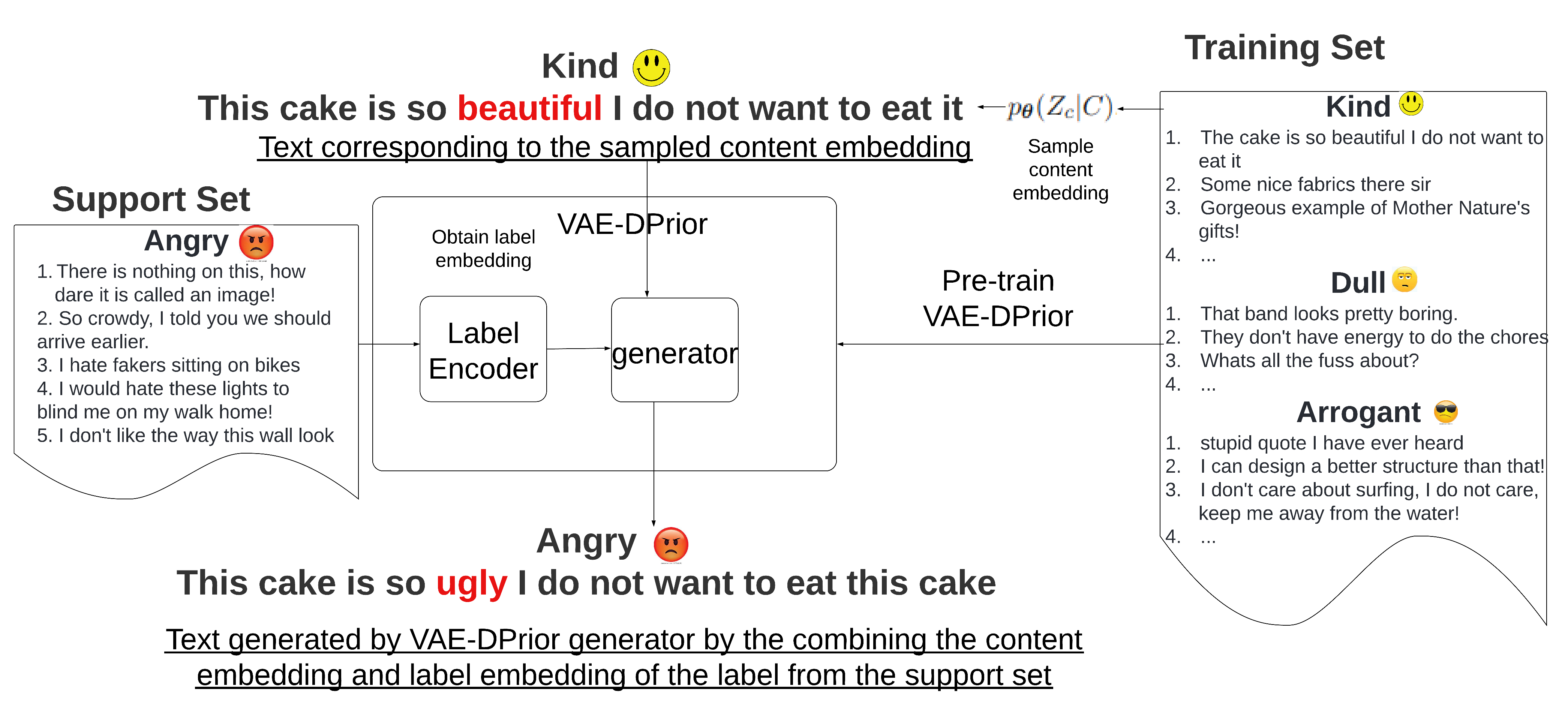}
    \caption{Generation of task-specific examples for data augmentation. In this example, the content representation is sampled from the training set, while the label representation is constructed based on the support set. 
    \vspace{-3mm}}
    \label{fig:example}
\end{figure}

Task-specific Natural Language Generation (NLG) aims to generate texts that satisfy desired attributes of target tasks, such as text style transfer~\cite{jin2020deep} and task-specific data augmentation~\cite{lee2021neural}. Herein, a task includes a set of task-specific labels, optionally a set of labeled texts for that task~\cite{han2020continual}. Although there is already a large amount of labeled data for various tasks, in many application scenarios, such as AI assistants for legal aid, the labeled data of new tasks are still difficult to acquire. As a result, there may be no or just a handful of labeled texts for target tasks. In such a low-resource setting, given a new task, it is desirable to i) identify which information in texts is task-specific and which is task-independent, and ii) systematically and consistently combine the label representations of the new task with task-independent content representations for text generation. As illustrated in Fig.~\ref{fig:example}, data augmentation needs to combine content representations from seen tasks with novel task labels. In contrast, text style transfer requires combining the content representations extracted from inputs with target styles. 

Most prior work assumes access to labeled data for supervised training. However, those models trained on seen tasks cannot generalize well to new tasks during inference~\cite{krishna2022few}. One of the key reasons is that the parameters of supervised models are tied to seen tasks such that a significant amount of fine-tuning data is needed for adapting to new tasks. For prompt-based and guided decoding methods~\cite{zhang2022survey}, although they require significantly less training data, it is still challenging to generate a large number of semantically diverse and coherent texts for new tasks in a robust way because they cannot well leverage the rich contents of the seen tasks.  

The key challenge of low-resource task-specific NLG is to disentangle content representations from label representations with few labeled data of target tasks. If content representations still contain task-specific information from seen tasks, they may well mislead the language generator after fusing with the representations of new tasks. Prior works tackle this problem by enforcing the random variables of content representations to be independent of those of label representations~\cite{cheng2020VIDisentangled}. However, in practice, both types of random variables are not always independent. For example, the random variables of emotion labels naturally depend on the contents of the events causing them.

In this work, we propose a deep VAE model with \textit{novel} disentanglement priors, coined \ourmodel, for task-specific natural language generation in the zero-shot and few-shot settings. In contrast to the widely used \textit{unconditional} priors in the VAE framework, the new priors are \textit{conditional}, satisfying a \textit{novel} property called $\epsilon$-disentangled, which motivates a new way of regularization for disentangling representations \textit{without forcing independence} between the corresponding random variables. The new priors build a constraint space for latent content representations and latent label representations with the aims to i) minimize information overlap between the two types of representations and ii) enable generalization across tasks with little labeled training data. 
One of the priors is a \textit{conditional} Gaussian mixture in the content subspace for sampling rich content representations without accessing original training data. Another type of priors is a \textit{conditional} multivariate Gaussian per label that associates latent label representations with task-specific information, requiring only a label name or a small set of labeled examples.  
Extending a pre-trained language decoder based on the prefix-tuning technique~\citep{li2021prefix} with those priors, our model is able to sample rich content representations of seen labels and combine them with the representations of new labels to generate \textit{diverse} and \textit{natural} sentences. In addition, we empirically observe that \ourmodel alleviates posterior collapse~\cite{wang2020posterior}, which is a long-standing problem of VAEs that makes it difficult to train a latent model to generate coherent and semantically diverse texts. 

To sum up, our key contributions are three-fold:
    i) We propose a \ourmodel model with \textit{novel} disentanglement priors for low-resource task-specific NLG tasks. It enables sampling diverse content representations directly from the content prior; 
    ii) We introduce $\epsilon$-disentangled, which sets a \textit{novel} regularization goal for disentangled representations;
    iii) Our model outperforms competitive baselines in the low-resource settings on the tasks of text style transfer and data augmentation for continual few/zero shot text classification. 

%


\section{Methodology}
\label{sec:method}


To tackle task-specific NLG tasks in low-resource settings, we introduce a deep generative model \ourmodel, which employs disentanglement priors, including a content prior for rich contents, to generate coherent and semantically diverse texts. We are provided with a large corpus of labeled sentences $\gD^{(0)} = \{\vx_i,y_i\}_{i=1}^n$ for an initial task $\mathcal{T}^{(0)}$, where a sentence $\vx_i \in \mathcal{X}$ is annotated with a seen label $y_i \in \gY$. The goal is to learn a single model that can generate diverse texts for any new task or a sequence of $K$ distinct new tasks \{$\mathcal{T}^{(1)}$, $\mathcal{T}^{(2)}$,...,$\mathcal{T}^{(K)}$\}. Each new task includes multiple novel labels, where a label $y \in \gY$ is associated with a label name and optionally a handful of example texts $\gD_{sup} = \{\vx_i,y_i\}_{i=1}^m$ as the \textit{support} set. The model is evaluated on both \textit{data augmentation for continual text classification} described in Sec. \ref{sec:exp} and \textit{few-shot text style transfer} detailed in Appendix \ref{sec:transfer_setting}.  

For evaluating data augmentation, a text classifier is trained \textit{sequentially} on $K$ new tasks and evaluated on the test sets of all seen tasks \{$\mathcal{T}^{(1)}$,...,$\mathcal{T}^{(t)}$\} till time $t$. For each task, its training data includes the texts generated by the NLG models in order to evaluate to what degree the augmented texts improve the classifier performance. In the zero-shot setting, the classifier is trained only on the generated texts using the label names of new tasks, while the support sets are also used for data augmentation and classifier training in the few-shot setting. 

\subsection{Theoretical Framework}
\label{sec:ivae}
\begin{figure}[t]
    \centering
    \includegraphics[width=\linewidth]{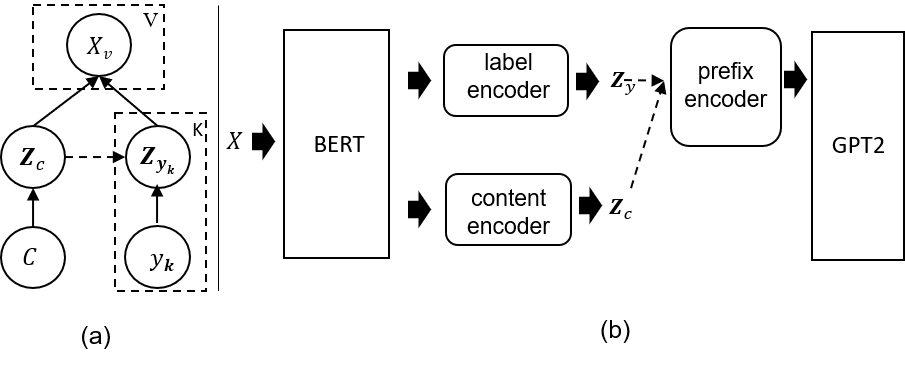}
    \caption{(a) A directed graphical model for disentanglement learning. (b) The architecture of \ourmodel.\vspace{-6mm}}
    \label{fig:DAG}
\end{figure}

In the absence of large training data for new tasks, one of the key challenges is to construct content representations and label representations in the latent space satisfying \textit{information purity}. As such, content representations should not contain label information, otherwise old task information in such content representations may contaminate combined representations for new tasks, vice versa.

Formally, the latent space is the sample space $\Omega$ for both content and label representations. In the corresponding probability space, we define a random variable vector $\rmZ_y$ for the latent representations of each label $y \in \mathcal{Y}$, a random variable vector $\rmZ_c$ for latent content representations. Then observable word sequences are denoted by the random variable vector $\rmX$, where each variable $X_v$ corresponds to a word in the vocabulary $\mathcal{V}$. The statistical dependencies between those random variables are illustrated by the Bayesian network in Fig.\ref{fig:DAG}(a), where $C$ denotes the prior knowledge of contents. The dashed arrow denotes a possible dependency between $\rmZ_y$ and $\rmZ_c$.


To achieve information purity, the learned models are expected to follow the structure illustrated in Fig.\ref{fig:DAG}(a) that there is no dependency between $C$ and $\rmZ_y$, as well as no dependency between $y$ and $\rmZ_c$. However, prior works on disentangled representation learning regularize the models by approximating $\rmZ_c \independent \rmZ_y$~\cite{cheng2020VIDisentangled,wang2021desiderata}, which may violate the true statistical relation between $\rmZ_c $ and $ \rmZ_y$. Even though $\rmZ_c \independent \rmZ_y$ holds after regularization, it does not imply $\rmZ_y \independent C$ and $\rmZ_c \independent y$. The random variable of a label can still depend on both $\rmZ_c $ and $ \rmZ_y$.

To address this limitation, we propose to regularize the priors of the latent variables for encouraging information purity. After training, we expect the mutual information of $I(\rmZ_y, Y)$ and $I(\rmZ_c, C)$ is high, while $I(\rmZ_y, C)$ and $I(\rmZ_c, y)$ is low or zero. One way to achieve this is that we make $I(\rmZ_y, y)$ high only in the dense regions of $\rmZ_y$ but force it to be low or zero in the dense regions of $\rmZ_c$, vice versa.  As a result, we expect little overlap between the dense regions of $\pTc(\rmZ_c|C)$ and those of  $\pTy(\rmZ_y|y)$. Then the distances between those priors are large. We characterize this property by introducing $\epsilon$-disentangled below.

\begin{definition}[$\epsilon$-disentangled]
\label{def:epsilson_disentangled}
Two distributions $\pTc(\rmZ_c | C)$ and $\pTy(\rmZ_y | y)$ are $\epsilon$-disentangled, if  $1/\diver_k(\pTc(\rmZ_c | C) || \pTy(\rmZ_y | y) ) \leq \epsilon$ and $\epsilon \in \mathbb{R}^+$, where $\diver_k$ denotes a divergence measure requiring no absolute continuity~\cite{royden1988real}, then $\pTc(\rmZ_c | C)$ and $\pTy(\rmZ_y | y)$ are $\epsilon$-disentangled w.r.t. the measure $\diver_k$.
\end{definition}


We refer to the priors satisfying $\epsilon$-disentangled as \textit{disentanglement priors}. In Appendix~\ref{sec:disentangled}, we conduct an in-depth discussion of this property. We show that if $\pTc(\rmZ_c | C)$ and $\pTy(\rmZ_y | y)$ are not $\epsilon$-disentangled under a mild assumption, at least one of them is non-identifiable, which is a leading cause of posterior collapse~\cite{wang2020posterior}.

\noindent\textbf{VAE with disentanglement priors.} Using the disentanglement priors, we employ the maximum likelihood principle for learning the parameters of the joint distribution $\prod_{y \in \gY}\pT(\rmX, \rmZ_c, \rmZ_y | C, y)$. The marginal distribution $\prod_{y \in \gY} \pT(\rmX | C, y) $ is given by 
\vspace{-2mm}

\begin{small}
\begin{equation}
\label{eq:joint_dist} 
   \int \prod_{y \in \gY}\pT(\rmX| \rmZ_c, \rmZ_y , y, C) \pTy(\rmZ_y | y) \pTc(\rmZ_c | C) d\rmZ_y d\rmZ_c .   
\end{equation}
\end{small}

We learn the above distribution in the VAE framework. Note that, the introduction of the conditions $C$ and $y$ makes the priors of both latent variables \textit{conditional}, which differs from vanilla VAEs that have only \textit{unconditional} priors for latent variables.

Given a dataset $\gD = \{(\vx_i, y_i)\}_{i=1}^n$, the training problem to learn the marginal distribution in Eq.~\ref{eq:joint_dist} is formulated as:

\vspace{-2mm}
\begin{small}
\begin{align}\nonumber
    \label{eq:opt_problem}
        \max & \sum_{i=1}^n \log p(\vx_i| C, y_i)\\
        & s.t.\hspace{2pt} \pTc(\rmZ_c | C) \text{ and } \pTy(\rmZ_y | y) \text{ are $\epsilon$-disentangled.}
\end{align}
\end{small}
\vspace{-2mm}

The disentanglement constraint is achieved by either carefully choosing priors satisfying $\epsilon$-disentangled, applying a divergence measure requiring no absolute continuity between priors as a regularizer, such as the Maximum Mean Discrepancy (MMD)~\cite{gretton2012mmd}, or both. In the following, we provide the model details and show how to derive an evidence lower bound (ELBO) in the VAE framework for this optimization problem. 



\subsection{Model Details}
\label{sec:model}


As illustrated in Fig.~\ref{fig:DAG}(b), the overall architecture consists of an inference module, a generator and priors. The inference module consists of a pre-trained \bert encoder, whose outputs serve as inputs of a label encoder and a content encoder, and a generator comprising a prefix encoder and a pre-trained \decoder with frozen parameters.

The VAE framework adopts variational distributions to approximate true distributions~\cite{kingma2019vaeBook}, which ends up maximizing an ELBO. We show in Appendix~\ref{sec:ELBO} 
that the ELBO objective takes the following form:

\vspace{-5mm}
\begin{align}
\label{eq:elbo}
    \begin{split}
        &\overbrace{\mathbb{E}_{q_{\vPhi}(\rmZ_c, \rmZ_y | \rmX, C, y)}[\log p_{\vTheta}(\rmX| \rmZ_c, \rmZ_y )]}^{\gL_r} \\
        &- \KL(\qP(\rmZ_c | \rmX, C)  \| \pTc(\rmZ_c | C))  \\ 
        &- \KL(\qP(\rmZ_y | \rmX, y)  \| \pTy(\rmZ_y | y)),
    \end{split}
\end{align}
where the first term is referred to as the reconstruction loss $\gL_r$, the other terms constitute regularizers. Following the convention of VAE, we refer to the network for $q_{\vPhi}(\rmZ_c, \rmZ_y | \rmX, C, y)$ as \textit{inference module}, the network for $p_{\vTheta}(\rmX | \rmZ_c, \rmZ_y)$ as \textit{generator}.

\noindent\textbf{Priors.} In the label subspace, we assume $\pTy(\rmZ_y | y)$ for a label $y$ is a simple factorized Gaussian distribution in form of $\mathcal{N}(\rmZ_y ; \vMu^p_y, \lambda_y\rmI)$, where $\lambda_y$ is a hyperparameter, its mean $\vMu^p_y$ is constructed by using the name embedding of label $y$ in the zero-shot setting, and by averaging the label name embedding and the embeddings of its support set examples in the few-shot setting. Each embedding is curated by feeding its word sequence to the label encoder shared with that of the inference module.

The content prior $\pTc(\rmZ_c | C)$ takes the form of $\sum_{k=1}^K \pT(M = k) \mathcal{N}(\rmZ_c ; \vMu^p_{c,k}, \lambda_c\rmI)$, where $M$ is the random variable indicating the membership to a component Gaussian. Inspired by neural topic modelling~\cite{wang2020neural}, we encode the prior knowledge of content $C$ into a $k$-means clusters, where we assume that there is a one-to-one correspondence between a component Gaussian and a cluster in the $k$-means clusters. The mean of a Gaussian component $\mathcal{N}(\rmZ_c ; \vMu^p_{c,k}, \lambda_c\rmI)$ is computed by $\rmW_c \vc_k$, a linear projection from the corresponding cluster centroid $\vc_k$. The $k$-means clusters are built from BERT sentence embeddings on the training data of seen tasks. Adding new topics is a matter of adding new clusters using incremental clustering techniques.

In Appendix \ref{sec:non-identifiability}, we show that $\pTc(\rmZ_c | C)$ and $\pTy(\rmZ_y | y)$ are $\epsilon$-disentangled with a small $\epsilon$ if their means are far from each other and their variances are sufficiently small.

\noindent\textbf{Inference Module.} The inference module is a \encoder~\citep{devlin2018bert} encoder augmented with an encoder for content, and an encoder for labels. Each encoder is built on top of the contextual embedding sequences produced by \encoder and yields latent representations of the target type. This design is not only parameter efficient but also leverages the strengths of a large-scale pre-trained transformer model. 

A \encoder model consists of multiple layers. To provide more model capacities to capture the differences between the two types of latent representations and preserve parameter efficiency, we freeze all layers of \encoder except the top most one so that the content encoder and the label encoder employ a top most transformer layer with different parameters respectively, while sharing all the remaining layers of \encoder.

In the label subspace, given a contextual word embedding sequence $\rmV_l = \{\vv_0, ..., \vv_u\}$ generated by the corresponding top-most layer of \encoder, the \textbf{LabelEncoder} implements $\qP(\rmZ_y |\rmX, y)$ in form of $\mathcal{N}(\rmZ_y ; \vMu^q_y, \text{diag}(\vSigma_y^2))$. In order to build a hidden representation focusing on label relevant information, we apply the label embedding $\vMu^p_y$ used in the label prior to $\rmV_y$ via soft attention. In particular, we compute an aggregated representation $\vh_y = \text{attention}(\vMu^p_y, \rmV_l)$ for a label $y$ by applying $\vMu^p_y$ as the query vector to attend all vectors of $\rmV_y$. We compute the mean $\vMu^q_y$ as a linear transformation of $\vh_y$ by using the weight matrix $\rmW^l_{\mu}$ and the logarithm of the variance $\log \bm{\sigma}_y$ as the linear transformation of another linear matrix $\rmW^l_{\sigma}$.

By applying the reparameterization trick~\citep{kingma2019vaeBook}, the latent label representation $\vz_y$ is a function of $\vMu^q_y$ and a stochastic noise.
The stochastic noise is added by the product of $\bm{\sigma}_y$ and the Gaussian noise $\bm{\epsilon}_y$ drawn from $\mathcal{N}(0,\rmI)$.  
\vspace{-0.5mm}
\begin{align}\nonumber
\label{eq:inf_label}
    \log \bm{\sigma}_y, \vMu^q_y &= \text{LabelEncoder}(\rmV_l)\\
    \vz_y &= \vMu^q_y + \bm{\sigma}_y \odot \bm{\epsilon}_y
\end{align}
\vspace{-1mm}
where $\odot$ denotes the element-wise product.

In the content subspace, we consider $\qP(\rmZ_c | \rmX, C) = \mathcal{N}(\rmZ_c ; \vMu^q_c, \text{diag}(\vSigma_c^2))$. Taking $\rmV_c$ from the corresponding top most layer of \encoder as input, \textbf{ContentEncoder} consists of a mean pooling layer followed by a linear layer for the mean and another linear layer for the logarithm of the variance of $\qP(\rmZ_c |\rmX, C)$. The same reparameterization trick is applied to obtain the latent representation $\vz_c$.

\noindent\textbf{Generator.} Given a pair of latent representations $(\vz_c, \vz_y)$, the generator captures $p(\rmX | \vz_c, \vz_y)$ factorized into the following autoregressive form. 
\vspace{-1mm}
\begin{align}
p_{\vtheta}(\vx|\vz_c, \vz_y) = \prod_{t=1}^{|\vx|}p_{\vtheta}(x_{t}|\vx_{<t},\vz_c, \vz_y) 
\end{align}

We employ a prefix-tuning technique~\citep{li2021prefix} that yields continuous prompts for the decoder in the low-resource situations. A continuous prompt is a continuous vector sequence of length $L$. The \textit{prefix encoder} consists of $L$ MLPs, each of which uses the architecture $\boldsymbol M_{\theta}[i,:] = \text{MLP}_{\theta}([\boldsymbol M'_{\theta}[i,:];\vz_y;\vz_c])$ for computing a vector at position $i$, where $\boldsymbol M'_{\theta} \in \mathbb{R}^{|M_{idx}| \times H'}$ is a learned matrix to encode the position information of the continuous prompt. In practice, we find it also useful to prepend the name embedding of the target label to the prefix. Further implementation details are available in Appendix~\ref{sec:implementation}.

In the low-resource settings, the mechanisms of constructing latent label and content representations across tasks should be consistent, otherwise labeled data needs to be provided for adjusting model parameters for alleviating those discrepancies. Therefore, we minimize parameters to update across tasks, use the same prior for content representations, and construct label embeddings using the same label encoder for different tasks. The parameters of the pre-trained encoder and the pre-trained decoder are frozen during training. Thus, we only need to train the parameters of the intermediate hidden layers between them on the data of initial tasks, which are also frozen for new tasks. Freezing parameters could effectively prevent the catastrophic forgetting of models when learning the new tasks.

\subsection{Training and Inference}
\label{sec:model_training}

Given a training corpus $\train = \{\vx_i, y_i\}_{i=1}^n$, we derive the objective function $\gL_{\vTheta, \vPhi}(\train) = \gL_r + \sum_y\gL_y(y) + \gL_{c}$ from the objective in Eq.~\ref{eq:opt_problem} and the ELBO, where $\gL_y$ and $\gL_{c}$ are the KL regularization terms from the ELBO. The constraint is removed by the usage of the disentanglement priors.

We first pre-train the whole model on the corpus of the initial task $\mathcal{T}^{(0)}$ without applying any disentanglement constraint and the regularizers derived from the ELBO, followed by fine tuning the model with all regularizers. In practice, we find out that the two-steps approach is important to achieve optimal empirical performance.


\noindent\textbf{Regularization in the Label Subspace.} The regularization term $\gL_y(y)$ in the label subspace is derived from $\KL(\qP(\rmZ_y | X, y)  \| \pTy(\rmZ_y | y))$. 

\vspace{-1mm}
\begin{small}
\begin{equation}
   \gL_y(y) = -\frac{1}{\lambda_y}\|\vz_y - \vMu_y^p\|^2 + \log \vSigma_y^q 
\end{equation}
\end{small}

The first term enforces latent label representations $\vz_y$ to be close to the label prototype $\vMu_y^p$ obtained from the label priors. In contrast, the corresponding regularization term in a vanilla VAE with unconditional Gaussian priors takes the form of $\|\vz_y\|^2$, which only makes the latent representations smooth without providing any label specific information. 

\noindent\textbf{Regularization in the Content Subspace.} Derived from $\KL(\qP(\rmZ_c | X, C)  \| \pTc(\rmZ_c | C))$, the regularization term $\gL_c$ takes the similar form as the loss of deep $k$-means~\cite{fard2020deepKmeans}. 
\begin{small}
\begin{align}
\begin{split}
         \gL_c = &\sum_{k=1}^K p(M = k|\vz_c)\big[-\frac{1}{2\lambda_c}\|\vz_c - \vMu_{c,k}^p\|^2 \big] + \log \vSigma_{c} 
\end{split}
\end{align}
\end{small}
We compute it by using EM. The term $\qP(M_k|\vx)$ is computed by the E-step. If soft-EM is considered, $\qP(M_k|\vx) = \frac{\exp(-\text{dist}(\vz_c, \vMu_{c,k}^p)/\tau)}{\sum_{k'}\exp(-\text{dist}(\vz_c,\vMu_{c,k'}^p)/\tau)}$, which denotes the probability of an example $\vx$ belonging to a cluster $k$ with a temperature $\tau$. In our experiments, we employ hard EM, where $\qP(M_k|\vx)$ indicates if the current Gaussian has the same the index as the one having the minimal Euclidean distance $\|\vz_c - \vMu_{c,k}^p\|^2$ among all components.

\noindent\textbf{Inference.} For data augmentation, our model samples a large number of texts from the model and filters out the ones that are not in accordance with the target labels. For each new label $y$, we construct the mean $\vMu^p_y$ of $\pTy(\rmZ_y| y)$ by averaging the embeddings of the label name phrase and optionally its associated texts from the support set. The corresponding embeddings are generated by feeding name phrases and texts into the label encoder. 
Then we sample a large number of content embeddings from the content prior $\pT(\rmZ_c | C)$. All combinations of label embeddings and content embeddings are fed to the generator to generate candidate examples. We find that the candidates of low quality are not in accordance with the target labels. Hence, we perform \textbf{quality control} by filtering out irrelevant ones. Specifically, we project each candidate to a latent representation using the label encoder, and rank all candidates w.r.t. the Euclidean distance between each representation and its associated name embedding. The top-$k$ candidates are taken as the final outputs.

\section{Experiments}
\label{sec:exp}
We evaluate our model on both continual few/zero shot text classification and few-shot text style transfer. The former requires sampling rich content representations from seen tasks, while the latter expects to retain task-independent contents from inputs. In both cases, it is desirable for models to systematically combine latent label and content representations across tasks in a consistent manner. 

The details of \textbf{few-shot text style transfer} are available in Appendix \ref{sec:exp_text_style_transfer}. We compare \ourmodel with five style-transfer baselines and show superior results on two datasets in the few-shot setting in terms of accuracy of style transfer, semantic relevance and naturalness of the generated text.

\subsection{Continual Zero/Few-shot Learning}



\noindent\textbf{Setting.} The general setting of continual zero/few-shot text classification has been introduced in Sec. \ref{sec:method}. 
Following a conventional continual learning setting~\citep{lopez2017gradient}, a memory $\mathcal{M}_{k}$ is associated with a task $\mathcal{T}^{(k)}$ to store a fixed number of training examples\footnote{The examples in a memory are selected either from the corresponding support set or augmented data.} per seen task. Upon the arrival of a new task, given the label names in the task and optionally a support set, a generative model produces new task-specific examples. A classifier is trained on a combined set of examples from the support set, the memories, and the augmented examples, and evaluated on the test sets of all seen tasks. 
The datasets we used are \empathe and \tacred. Please refer to Appendix ~\ref{sec:cl_setting} and ~\ref{sec:cl_aug} for more details.

\noindent\textbf{Evaluation.} We use the widely adopted metric $\text{ACC}_{\text{avg}}$ in continual learning, which measures the performance by averaging the accuracies of the classifier on test sets of all seen tasks $\{\mathcal{D}_{test}^{(1)}$,...,$\mathcal{D}_{test}^{(k)}\}$, namely $\text{ACC}_{\text{avg}} = \frac{1}{k} \sum^{k}_{i=1} acc_{i}$~\citep{lopez2017gradient}. In addition, to measure the diversity of generated examples, we calculate the average similarity scores between all pairs of examples within each label, \textit{i.e.} $\frac{1}{|\mathcal{Y}|}\sum_{i,j} sim(\vx_i, \vx_j) \mathbbm{1}[ \argmax p(y | \vx_i) == \argmax p(y | \vx_j) ]$, where we use BLEU~\citep{papineni2002bleu} and word mover distance (WMD)~\citep{kusner2015word} as the similarity functions. The \textit{lower} scores indicate more diversified examples within each label.


\noindent\textbf{Baselines.} We compare five data augmentation baselines: \textit{i)} \eda~\citep{wei2019eda} randomly deletes, substitutes, inserts or swaps words in the original sentences. \textit{ii)} \cbert~\citep{ma2019nlpaug} uses BERT to determine the position to insert or substitute words. \textit{iii)} \back~\citep{sennrich2015improving} augments datasets by generating the paraphrases of the original sentences through round-trip translation, \textit{iv)} \lamb~\citep{kumar2020data} trains a GPT2 to generate examples conditioned on the label text and uses a classifier to filter out low-quality examples as in our work. \textit{v)} \ex~\cite{lee2021neural} applies T5~\citep{raffel2020exploring} and extrapolation technique to increase the diversity of generated examples and 
deal with the low-resource setting. \textit{vi)} \optimus~\citep{li2020optimus} is our backbone model which is in an auto-encoder framework that uses BERT as the encoder and GPT2~\citep{radford2019language} as the decoder. \textit{vii)} \casuallens~\citep{hu2021causal} improves the training of \optimus  using an intervention and a counterfactual losses. Both \optimus and \casuallens are designed for controllable text generation. We use them for data augmentation by assessing their ability for label-conditional generation. 


\noindent\textbf{Main Results and Discussions.}
\begin{table}[t]
\centering
  \resizebox{0.9\textwidth}{!}{%
  \begin{tabular}{|c||ccc|ccc|}
    \toprule
    \multirow{2}{*}{Methods} &
      \multicolumn{3}{c|}{\tacred} &
      \multicolumn{3}{c|}{\empathe}\\
      & 0  & 1 & 5  & 0 & 1  & 5  \\
      \midrule \midrule
    \noaug & 11.21 & 22.02 &  36.87  & 9.75  & 14.20  & 24.07  \\
\hline
    \eda & - &20.64 & 32.83 & -  & 13.72& 20.16   \\
    \cbert & - & 20.21 & 35.43 &- & 13.82 & 21.52 \\
    \back & - & 24.23 & 33.60 &-  & 14.06 & 20.37  \\
\hline
\lamb & 16.23 & 20.16 & 32.14 &13.39  & 14.45 & 21.95 \\
\ex & 15.57 & 19.83  & 32.87 &11.91 &16.82 &24.75  \\
\hline
\optimus & 17.12 & 19.99  & 28.77 &9.93 &14.32 &18.21  \\
\casuallens & 9.83 & 17.17  & 25.76 & 9.72 & 11.72 & 16.49  \\
\hline
\ourmodel & \textbf{31.34} & \textbf{37.31} & \textbf{44.17} & \textbf{18.08} & \textbf{22.71} & \textbf{31.82}\\
                
    \bottomrule
  \end{tabular}%
  }
    \caption{The $\text{ACC}_{\text{avg}}$ of the classifier across the tasks with different data augmentation methods. '-' indicates that zero-shot is not applicable to the corresponding augmentation methods.     \vspace{-3mm}}
  \label{tab:cl_aug}

\end{table}
%
We compare first the baselines with our model in its best setting, coined \ourmodel, which applies both the disentanglement priors and the MMD regularizer between the priors. The results in Table~\ref{tab:cl_aug} show that it outperforms all data augmentation baselines on all zero/few-shot learning settings by significant margins. The augmentation approaches such as \eda, \cbert and \back generate adversarial examples of the original sentences via manipulation of words or paraphrasing. However, adversarial distributions are not the same as the true distribution, thus their generated examples do not improve the continual learning performance. They even degrade the performance in the five-shot setting in comparison to that without data augmentation. 

Although \lamb, \ex, \optimus and \casuallens aim to learn the true distribution from labeled data, we observe that they often fail to generate texts in accordance with correct labels, especially for new tasks. Thus, their performance cannot be improved given more labeled examples of new tasks. In contrast, \ourmodel achieves a significantly higher degree of compositional generalization across tasks, evident by high average accuracy of the classifier trained on its generated examples. The performance of the classifier further improves when our model is fed with more labeled examples of new tasks.  

\begin{table*}[ht]
    \vspace{-3mm}
\centering
  \resizebox{0.85\textwidth}{!}{%
  \begin{tabular}{|cc||cccccccc|}
    \toprule
   
      Datasets & Metrics & \eda  & \cbert & \back  & \lamb & \ex & \optimus & \casuallens &\ourmodel  \\
       \midrule \midrule
    \multirow{2}{*}{\tacred} & BLEU$\downarrow$ & 38.24 & 28.12 &  96.83  & 95.85  & 26.61 & 47.04 & 18.78 &\textbf{4.14}  \\
    & WMD$\downarrow$ & 97.91 & 96.83 &  99.93  & 99.28  & \textbf{83.24}  & 94.94 & 88.24 &86.08  \\
\hline
    \multirow{2}{*}{\empathe} & BLEU$\downarrow$ & 31.64 &24.97 & 96.79 & 74.82  & 44.55 & 55
    12 & 9.46 &\textbf{5.57}   \\
    & WMD$\downarrow$ & 97.89 & 96.85 & 99.92 & 97.20  & 94.57 & 98.15 & \textbf{79.10} & 92.67   \\
                
    \bottomrule
  \end{tabular}%
  }\vspace{-2mm}
    \caption{The diversity scores of the generated examples measured with BLEU and WMD on one-shot learning.\vspace{-3mm}}
  \label{tab:cl_diversity}
    \vspace{-1mm}
\end{table*}

We also evaluated \textit{diversity} of the generated examples of all augmentation methods in the one-shot setting, presented in Table~\ref{tab:cl_diversity}. The generated sentences from \eda, \cbert and \back are mostly paraphrases of the original sentences. Therefore, they cannot significantly diversify examples at the semantic level. \lamb generates data examples conditioning on the label names and the first $k$ words of the original sentence, which also lacks diversity. \ex enriches the diversity by extrapolating the novel samples from the existing sentences within the novel labels. \optimus and \casuallens employ a GAN~\citep{goodfellow2014generative} and a conditional GAN~\citep{mirza2014conditional} respectively to generate diversified latent vectors for the generation of examples with the novel labels. However, with merely one or five sentences per label, such methods only generate a small sample of texts with novel labels. In contrast, \ourmodel can combine plenty of seen content representations acquired from the past with representations of new labels to generate high-quality sentences.

\subsection{Ablation Study}

\begin{figure*}
  \resizebox{0.85\textwidth}{!}{%
    \vspace{-1mm}
     \centering
     \begin{subfigure}[b]{0.3\linewidth}
         \centering
         \includegraphics[width=\linewidth]{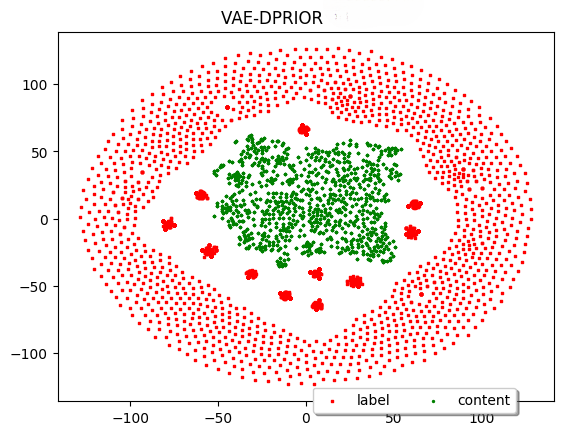}
     \end{subfigure}
     \hfill
     \begin{subfigure}[b]{0.3\linewidth}
         \centering
         \includegraphics[width=\linewidth]{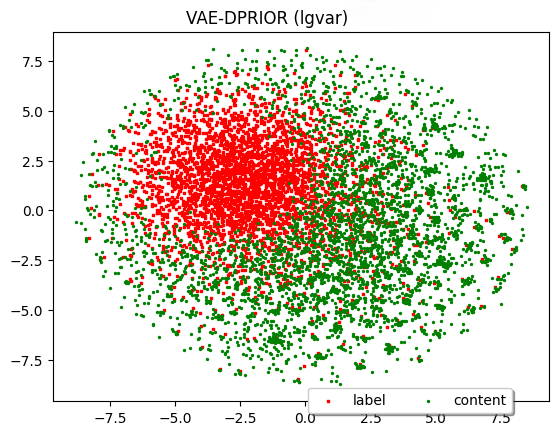}
     \end{subfigure}
     \hfill
          \begin{subfigure}[b]{0.3\linewidth}
         \centering
         \includegraphics[width=\linewidth]{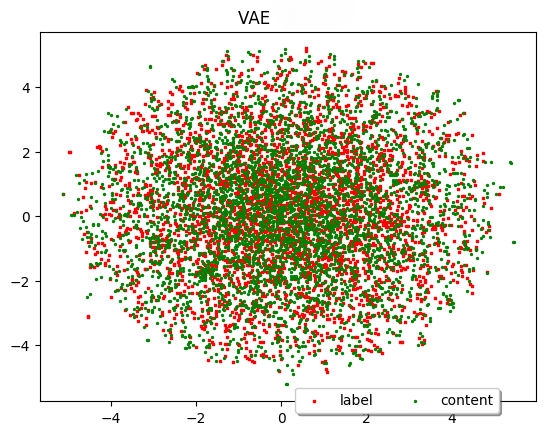}
     \end{subfigure}
     \hfill
     }
        \caption{The label (red) and content (green) representations sampled from label and content priors of \ourmodel, \ivaelgvar and \vae trained on \tacred.    \vspace{-3mm}}
        \label{fig:rep_picture}
            \vspace{-3mm}
            
\end{figure*}

\noindent\textbf{Disentanglement.} To show the importance of $\epsilon$-disentanglement, we remove the constraint of the optimization problem \eqref{eq:opt_problem} by using only the pre-trained model resulted from the first training step, denoted by \auto. As shown in Table~\ref{tab:abl_vae_div}, it suffers from a significant drop in terms of all metrics in the one-shot setting. In the same table, we also report the comparisons with alternative priors: (i) unconditional priors as in the vanilla VAE (\vae), (ii) the same priors but increasing the variance coefficients of the two priors from 1 to 50 (\ivaelgvar), (iii) a Gaussian mixture with randomly initialized means as the content prior but do not fine-tune the parameters of the prior (\ivaerand), (iv) same as (iii) but fine-tune the parameters of the content prior (\ivaerandft), and (v) a simple factorized Gaussian conditioned on the averaged sentence embedding of all sentences of initial tasks as the content prior (\ivaegauss). In the one-shot setting, the accuracy drops by more than 7\% and 3\% on \tacred and \empathe, respectively, using the alternative or no priors, indicating the importance of $\epsilon$-disentanglement. Increasing the variance of our priors also jeopardize the $\epsilon$-disentanglement. As evident in Fig.~\ref{fig:rep_picture} using the t-SNE~\cite{van2008visualizing}, it is clear that the priors of \vae and \ivaelgvar are severely overlapped in contrast to \ourmodel. 

We further investigate how the disentanglement regularizers influence our model by removing MMD or replacing MMD with GAN, HSIC, and IDEL~\citep{cheng2020VIDisentangled}. As in Table~\ref{tab:abl_dis}, except for MMD, the other disentanglement regularizers bring almost no improvement to \ourmodel. HSIC, GAN and IDEL enforce independence between latent variables but even hurt the performance. We observe that the GAN-based regularizer causes mode collapse, because \ourmodel with GAN tends to generate overly similar examples. In contrast, if we apply the MMD to the other types of VAEs, such as a vanilla VAE, they lead to improved performance (see Appendix~\ref{sec:mmd_vae}) because the other VAEs do not have the ability to disentangle representations. 
\begin{table}[t]
\centering
  \resizebox{0.9\textwidth}{!}{%
  \begin{tabular}{|c||ccc|ccc|}
    \toprule
    \multirow{2}{*}{Methods} &
      \multicolumn{3}{c|}{\tacred} &
      \multicolumn{3}{c|}{\empathe}\\
      & 0  & 1 & 5  & 0 & 1  & 5  \\
      \midrule  \midrule
    \ourmodel & 31.34 & 37.31 & 44.17 & 18.08 & 22.71 & 31.82 \\
\hline
- \mmd & 29.21 & 37.18 & 43.34 &17.79& 22.43 & 32.27\\
\hline
    -/+ \adv & 25.72 & 27.97 & 35.98 &14.41 & 17.65 & 24.17 \\
    
    -/+ \hsic & 8.46 & 14.09 & 42.25 &18.02  & 21.39 & 32.28  \\
    -/+ \idel & 29.79 & 36.10 & 43.08 &16.12  & 22.00 & 32.58  \\
    \bottomrule
  \end{tabular}%
  }
    \caption{ The $\text{ACC}_{\text{avg}}$ of \ourmodel with different disentanglement losses in zero/few-shot learning.
     \vspace{-3mm} }
  \label{tab:abl_dis}
    \vspace{-3mm}
\end{table}



\noindent\textbf{Posterior Collapse.}
\begin{table}[t]
    \vspace{-1mm}
\centering
  \resizebox{0.9\textwidth}{!}{%
  \begin{tabular}{|c||ccc|ccc|}
    \toprule
    \multirow{2}{*}{Methods} &
      \multicolumn{3}{c|}{\tacred} &
      \multicolumn{3}{c|}{\empathe}\\
      &  $\text{ACC}_{\text{avg}}\uparrow$ & BLEU$\downarrow$ & WMD$\downarrow$  & $\text{ACC}_{\text{avg}}\uparrow$ & BLEU$\downarrow$  & WMD$\downarrow$  \\
      \midrule  \midrule
    \ourmodel &  37.31 & 4.14 & 86.08 & 22.71 & 5.57 & 92.67  \\
\hline
    -/+ \auto &  30.80 & 4.76 & 88.80 & 17.18 & 7.11 & 95.90 \\
\hline
     -/+ \ivaegauss & 13.74 & 6.16 & 85.25 & 13.28 & 14.31 & 92.78 \\
         -/+ \ivaelgvar & 23.24 & 2.83& 57.44 &  19.22 & 20.74& 94.69 \\
        -/+  \ivaerand  & 20.41 & 75.46 & 97.53 & 12.60 & 23.96 & 93.84 \\       
        -/+  \ivaerandft  & 23.79& 87.31  & 98.87 & 15.30 & 88.60 & 99.31 \\    
\hline
    -/+ \vae & 19.23&46.80&95.19 & 17.03&54.34&97.04 \\
    -/+ \vqvae & 27.84 &13.27& 88.78 & 10.14&11.41&87.29\\
    -/+  \cvae & 13.22&2.97&59.69 & 13.15&5.26&76.87 \\     
           -/+  \vampvae  & 19.08 &30.47& 82.31 & 17.47&41.00&95.75 \\    
    \bottomrule
  \end{tabular}%
  }
    \caption{ The $\text{ACC}_{\text{avg}}$ and diversity scores of the models with different VAE frameworks on one-shot learning. Sample from prior label and content distributions.
     \vspace{-3mm} }
  \label{tab:abl_vae_div}
    \vspace{-3mm}
\end{table}
Classical VAEs, such as vanilla VAE (\vae), suffer from a notorious problem called \textit{posterior collapse}. Those models learn non-injective mappings between latent variable values and the likelihoods; thus many latent representations are mapped to the same model outputs. We investigate this problem by comparing \ourmodel with \vae, \vqvae~\citep{oord2017neural}, \cvae~\citep{jang2016categorical}, \vampvae~\citep{tomczak2018vae} and \ourmodel with alternative priors described before in terms of diversity and accuracy. If there is severe posterior collapse, models will generate similar texts indicated by the high BLEU scores, and the classifier trained on the augmented data would perform poorly. Unsurprisingly, the results in Table~\ref{tab:abl_vae_div} show that \ourmodel largely outperforms those VAEs. Although \vampvae also introduces conditional priors, the latent variables of its priors do not require to be $\epsilon$-disentangled with a small $\epsilon$. The \ivaerandft even generate almost identical texts. 



Posterior collapse should lead to high ratios of duplicated outputs. Thus we feed each model with 200 diverse latent variable values randomly sampled from their priors and compute the duplicate ratios per label. \ivaerandft has the highest ratio 97.38\%, followed by \vae, \vampvae, \ivaerandft, \cvae and \vqvae with a duplicity ratio of 78.33\%, 70.38\%, 8.06\%, 6.10\% and 3.09\%, respectively, on \empathe in the one-shot learning.  In contrast, our model generates no duplicates with those latent variable values.

We also investigate the quality of the outputs of those models by sampling representations from the posterior distributions, $\qP(\rmZ_y | X, y)$ and $\qP(\rmZ_c | X, C)$. The duplicate ratio of \vae drops to merely 4.27\%, while that of \vampvae increases to 97.95\% on \empathe. Our model still achieves a zero duplicate ratio. However, appendix~\ref{app:inference_post} shows that the models sampling from posteriors achieve comparable results as those sampling from priors in terms of accuracy. Therefore, sampling contents from the posteriors may reduce duplicate ratios for some of the models but their generated examples still cannot have comparable quality as our model. 

\noindent\textbf{Quality Control.}
\begin{table}[t]
    \vspace{-2mm}
\centering
  \resizebox{0.9\textwidth}{!}{%
  \begin{tabular}{|c||ccc|ccc|}
    \toprule
    \multirow{2}{*}{Methods} &
      \multicolumn{3}{c|}{\tacred} &
      \multicolumn{3}{c|}{\empathe}\\
      &  $\text{ACC}_{\text{avg}}\uparrow$ & BLEU$\downarrow$ & WMD$\downarrow$  & $\text{ACC}_{\text{avg}}\uparrow$ & BLEU$\downarrow$  & WMD$\downarrow$  \\
      \midrule  \midrule
    + \lencoder &  37.31 & 4.14 & 86.08 & 22.71 & 5.57 & 92.67\\
\hline
    -/+ \random &  24.21 & 3.81 & 80.75 & 21.03 & 4.64 & 85.68 \\
    -/+ \perplexity & 23.32 &4.72 & 80.98  & 16.59 & 4.81 & 73.94 \\
    -/+ \bert &  33.21 &3.96 & 81.81 & 20.40 & 7.81 & 75.77\\
    \bottomrule
  \end{tabular}%
  }
    \caption{ The $\text{ACC}_{\text{avg}}$ and diversity scores of \ourmodel with different quality control methods on one-shot learning.    \vspace{-3mm}}
  \label{tab:abl_quality}
      \vspace{-3mm}
\end{table}

We compare our inference method (+\lencoder) with three alternative methods: \textit{i)} only use a pre-trained BERT to encode each output text into an embedding with mean pooling and compare it with the average BERT embeddings of labels and support sets ( +\bert); \textit{ii)} select $k$ examples with the lowest perplexity calculated by GPT2 ( +\perplexity); and \textit{iii)} randomly select $k$ outputs ( + \random). Table \ref{tab:abl_quality} shows that different methods indeed influence the final performance of the data augmentation. Although the BLEU and WMD metrics show that baseline filtering methods all increase the diversity, the quality of selected examples is actually lower. But even with the worst performing filtering method, ( +\perplexity), our method can still outperform other baselines on both datasets. We observe that BERT with the encoder trained with label condition prior performs much better than only using the backbone model, ( +\bert), in terms of selecting high quality examples, proving that the label prior condition could help the encoder generalize well on the novel labels.












\section{Related Work}

\noindent\textbf{Variational Autoencoder.}
A large series of work learn representations based on generative models, such as Variational Autoencoder (VAE)~\citep{kingma2019vaeBook}. 
A standard VAE minimizes the Kullback–Leibler divergence between the parametric posterior and the true posterior. Different posterior integrates various properties to the generative models, VQ-VAE~\citep{oord2017neural} parameterizes a posterior distribution of discrete latent random variables, Categorical-VAE~\citep{jang2016categorical} and Vamp-VAE~\citep{tomczak2018vae} constructs mixture of posteriors on learnable pseudo-inputs. They are not capable of remembering rich contents in NLG. Gaussian Mixture VAE~\citep{dilokthanakul2016deep} uses the mixture of Gaussian as the prior as well. It is not designed for disentanglement as it uses only one prior. 
~An Identifiable Double VAE~\citep{mita2021identifiable} does not use two different priors for different subspace. It uses only one prior based on observed random variables to remove the observed information from the latent representations in order to achieve disentanglement. In another word, it has no component to remember contents for continuous few shot setting.
Our work considers the posterior on multiple types of disentangle random variables, which is potentially of more expressiveness.

\noindent\textbf{Disentangled Representation Learning.} Disentanglement of representations is one of the ultimate goals of deep learning. The existing methods are either unsupervised or supervised~\cite{higgins2018towards}. The unsupervised ones mainly fall into either the framework of VAE~\cite{burgess2018understandingBetaVAE} or Generative Adversarial Learing (GAN)~\cite{tran2017disentangledGan}. The recent works have also incorporate causality theories for robustness~\cite{hu2021causal}. There are growing interests in applying disentangled representation learning in NLP applications, such as text style transfer~\cite{john2018disentangledRepTextStyle} and mitigating gender bias~\cite{liu2020mitigatingGenderBias}. However, it is challenging for those  NLP approaches to work in the low-resource settings because they do not store rich content information inside models~\cite{romanov2018adversarial}. 


\noindent\textbf{Controllable Text Generation.}
Our method decomposes content and (attribute) label, where the label could be considered as additional control signal for text generation. We connect our work to those text style transfer (TST) and controllable text generation (CTG). 
\textit{Representation disentanglement} is an important line of research in TST, which disentangles content and attribute representations \citep{john2018disentangledRepTextStyle}. 
Many disentanglement approaches are proposed to minimize the dependence between these two representations, such as mutual information \citep{yuan2020improving} and orthogonality \citep{wei2021orthogonal}. CTG controls the text generation of language models by 
smart prompt design~\cite{li2021prefix,shin2020autoprompt} or training conditioned on the controllable variables~\cite{li2020optimus,hu2021causal}. Our work is highly aligned with~\cite{li2020optimus,li2021prefix}. Since \cite{li2020optimus} and \cite{li2021prefix} have similar implementations, and \cite{li2020optimus} was designed for generation conditioned on latent variables, we pick \cite{li2020optimus} as one of our baselines.


\section{Conclusion}
In this work, we propose a VAE model with disentanglement priors for disentangled representation learning in low resource controllable NLG tasks. The disentanglement priors satisfy a novel property called $\epsilon$-disentangled which builds a constraint space for the training problem. This model is able to effectively combine rich content representations sampled from a conditional content prior and task-specific representations for new tasks. Its empirical performance outperforms the baselines on continual zero-shot/few-shot text classification and few-shot text style transfer by a wide margin.

\section*{Limitations}
We have studied $\epsilon$-disentangled only in the VAE framework for task-specific language generation, though we believe it should be useful for a wide range of latent models. Although the content prior of our model can already be used to sample rich content representations, there is a possibility to store more information and represent a even richer content space reflected in real-world scenarios. In addition, our model has not considered application scenarios with limited computing resources. Though it is beyond the scope of this work, due to the heavy use of pre-trained large-scale language models, the deployment of our model in those cases is particularly challenging. 
\section*{Acknowledgement}
This material is based on research sponsored by Air Force Research Laboratory and DARPA under agreement numbers FA8750-19-2-0501 and HR001122C0029. The U.S. Government is authorized to reproduce and distribute reprints for Governmental purposes notwithstanding any copyright notation thereon. The computational resources of this work are supported by the Multi-modal Australian ScienceS Imaging and Visualisation Environment (MASSIVE).
\bibliography{anthology,custom}
\bibliographystyle{acl_natbib}

\newpage
\appendix
\onecolumn
%
\section{Implementation Details}
\label{sec:implementation}
We use learning rate of 5e-5 for our method. The training epochs for our generation model in continual few-shot learning are 120 and 160 for \empathe and \tacred, respectively. All experiments are run five times with different random seeds and we report the average accuracies. The number of clusters for the deep content clustering loss are 1600, 800 and 3200 when training model on \empathe, \tacred and \persona, respectively. All the methods are trained on the V100 GPUs. The number of total parameters is 455864068 and the total number of total trainable parameters is 401751808. For style transfer, the training epochs are 120 and 160 for \empathe and \persona, respectively. We use BERT-small~\cite{turc2019well} as the backbone of the label and content encoders and GPT2-medium as the decoder. For data augmentation in continual few-shot learning, each label is augmented with 50 examples generated by different augmentation methods. \optimus is not designed for style transfer. We adapt it to conduct style transfer by prepending a label phrase as a prompt before the input sentence. Style transfer can be done by altering the current label phrase to novel labels in the new tasks.
\section{Continual Few-shot Learning}
\subsection{Setting} 
\label{sec:cl_setting}
We consider a continual few-shot learning setting similar as~\citep{antoniou2020defining}. 
The text classification model $\pi^{c}_{\theta}: \mathcal{X} \rightarrow \mathcal{Y}$ is trained sequentially on $K$ distinct tasks \{$\mathcal{T}^{(1)}$, $\mathcal{T}^{(2)}$,...,$\mathcal{T}^{(K)}$\}. The initial task $\mathcal{T}^{(1)}$ includes the training and test set $(\mathcal{D}_{train}^{(1)},\mathcal{D}_{test}^{(1)})$ while the succeed tasks $\mathcal{T}^{(k \succ 1)}$ includes the support and test sets $(\mathcal{D}_{sup}^{(k)},\mathcal{D}_{test}^{(k)})$, where we assume $\mathcal{D}_{train}^{(1)}$ includes enough training data for each base class while $\mathcal{D}_{sup}^{(k)}=\{\vx,\vy\}_{i=1}^{N\times |C_k|}$ includes only $N$-shot instances per new class. The classes on $\mathcal{T}^{(k)}$ are disjoint from classes of previous tasks, $C_{1:k-1}\cap C_k = \emptyset$. As a conventional continual learning setting as in~\citep{lopez2017gradient}, a memory $\mathcal{M}_{k}$ is associated with  $\mathcal{T}^{(k)}$ to store a fixed number of training instances (either examples selected from the support sets or the synthetic data) per each seen task.   

Upon arrival of each task, the classifier $\pi^{c}_{\theta}$ is trained on a combined set of instances from the support set, memory set, and the augmented examples generated by our generative model. 
To generate augmented examples, we sample content from the fixed clustering obtained using the large training data in $\mathcal{T}^{(1)}$, and sample labels from $C_{1:k}$. Note that for our model, as long as we have the generative model and store the label embeddings, we could regenerate examples from all old tasks. Therefore, the memory is not necessary for our model. But for a fair comparison with other baselines, we still assume there is fixed memory for each old task and use only the examples from this memory to replay the classifier.
%
We apply our data augmentation method to EMAR~\citep{han2020continual}, a  SOTA continual learning approach for text classification. We follow EMAR ~\citep{han2020continual} for the classifier architecture and how to update its parameters in continual learning.
\subsection{Datasets} 
\label{sec:cl_aug}
\tacred is a relation detection dataset which includes 42 relations. Following the settings in~\cite{wang2019sentence,han2020continual}, examples are clustered into ten groups given the word embeddings of the label phrases. 5 groups are randomly selected as the initial task. The support and test set from each task are drawn from each of the rest tasks. We randomly generate the support sets five times with different random seeds as well. Each support set includes 0, 1, or 5 examples.

\subsection{Influence of MMD on VAEs}
\label{sec:mmd_vae}
Table \ref{tab:dis_comparision} shows the performance of VAEs on five-shot learning of \tacred with or without MMD. We present only five-shot results as we found that the MMD brings almost no improvement to different VAEs on zero/one-shot learning. But it consistently leads to performance improvement on all VAEs except for \ourmodel when the number of shots increases. We conjecture that disentanglement regularization perform better when there is enough label-specific information.
\begin{table}[ht]
\centering
  \resizebox{0.9\textwidth}{!}{%
  \begin{tabular}{|c||ccccc|}
    \toprule
   
      $\text{ACC}_{\text{avg}}$  & \ivae & \vae & \vqvae & \cvae &\vampvae  \\
       \midrule \midrule
    w/o \mmd  & 43.35 &  27.86& 35.27 & 30.77 & 34.38  \\
    w/ \mmd & 43.13 &  29.72& 36.27 & 33.32 & 35.91 \\
                
    \bottomrule
  \end{tabular}%
  }
    \caption{The $\text{ACC}_{\text{avg}}$ of \ourmodel and other VAEs with or without \mmd of five-shot learning on \tacred. The representations are sampled from the label and latent representations from the posterior distributions. \vspace{-3mm}}
  \label{tab:dis_comparision}
    
\vspace{-3mm}
\end{table}

\subsection{Influence of Sampling from Posterior Distribution}
\label{app:inference_post}
Table~\ref{tab:abl_vae_div_post} shows the performance of the classifiers using augmented data sampled from posterior distributions of VAEs. Since sampling representations from posterior distributions, $\qP(\rmZ_y | X, y)$ and $\qP(\rmZ_c | X, C)$, requires text $X$ as the input, we feed all the text from the training sets in previous tasks to the content encoder to obtain the content representations and the text in the support set of new tasks to the label encoder to obtain the label representations. We combine the two types of representations to get augmented data for new tasks. Notice that the \auto setting in Table~\ref{tab:abl_vae_div} adopts a similiar way to sample examples for data augmentation except that the content and label representations $\vz_c$ and $\vz_y$ are generated by using \textbf{LabelEncoder} and \textbf{ContentEncoder} directly.
\begin{table}[t]
\centering
  \resizebox{0.9\textwidth}{!}{%
  \begin{tabular}{|c||ccc|ccc|}
    \toprule
    \multirow{2}{*}{Methods} &
      \multicolumn{3}{c|}{\tacred} &
      \multicolumn{3}{c|}{\empathe}\\
      &  $\text{ACC}_{\text{avg}}\uparrow$ & BLEU$\downarrow$ & WMD$\downarrow$  & $\text{ACC}_{\text{avg}}\uparrow$ & BLEU$\downarrow$  & WMD$\downarrow$  \\
      \midrule  \midrule
    \ourmodel &  37.11 &4.06& 86.64 & 28.40&5.10&88.74  \\
     -/+ \ivaegauss & 20.21 &57.73& 96.83 & 15.67&64.27&98.10 \\
        -/+  \ivaerand  & 26.13 &68.69& 97.02 & 14.20&100.00&100.00 \\       
        -/+  \ivaerandft  & 25.82 &40.49& 94.16 & 17.31&50.49&96.93 \\    
\hline
    -/+ \vae & 27.06&11.78&91.30 & 18.60 & 20.74 & 94.69 \\
    -/+ \vqvae & 28.12 &12.68& 88.28 & 11.85&10.03&86.27\\
    -/+  \cvae & 16.82&5.49&69.28 & 13.80&10.31&89.51 \\     
           -/+  \vampvae  & 17.46&100&100 & 16.58&98.19&99.92 \\    
    \bottomrule
  \end{tabular}%
  }
    \caption{ The $\text{ACC}_{\text{avg}}$ and diversity scores of the models with different VAE frameworks on one-shot learning. The representations are sampled from the posterior label and content distributions.
     \vspace{-3mm} }
  \label{tab:abl_vae_div_post}
    \vspace{-3mm}
\end{table}

\subsection{Accuracies of \ourmodel on \persona and \fewrel}
Table \ref{tab:acc_fewrel_persona} shows the performance of the baselines and \ourmodel on one-shot learning of \persona and \fewrel. Please notice that we use the exact same \fewrel dataset as in~\cite{wang2019sentence,han2020continual} except that we split the tasks in a different way. \ourmodel performs best on both datasets as well. 
\begin{table}[ht]
\centering
  \resizebox{\textwidth}{!}{%
  \begin{tabular}{|c|ccc|ccc|cc|c|}
    \toprule
   
      $\text{ACC}_{\text{avg}}$  & \noaug  & \eda & \cbert  & \back & \lamb & \ex & \optimus &\casuallens & \ourmodel \\
       \midrule \midrule
    \persona  & 8.65 & 10.22 &  10.52  & 11.40 &  10.58& 12.51 & 9.15 & 7.74 & 15.48  \\
    \fewrel & 42.64 & 46.88 &  53.93  & 46.45 &  46.71& 45.09 & 34.08 & 19.82 & 71.81  \\
                
    \bottomrule
  \end{tabular}%
  }
    \caption{The $\text{ACC}_{\text{avg}}$ of the baselines and \ourmodel in one-shot learning on \persona and \fewrel. \vspace{-3mm}}
  \label{tab:acc_fewrel_persona}
    
\vspace{-3mm}
\end{table}


\subsection{Accuracies of \ourmodel using Soft and Hard EM}
\begin{table}[ht]
\centering
  \resizebox{0.4\textwidth}{!}{%
  \begin{tabular}{|c|cc|}
    \toprule
   
      $\text{ACC}_{\text{avg}}$  & \tacred  & \empathe \\
       \midrule \midrule
    Hard EM  & 37.31 & 22.71   \\
    Soft EM & 39.47 & 25.43 \\
                
    \bottomrule
  \end{tabular}%
  }
    \caption{The $\text{ACC}_{\text{avg}}$ of \ourmodel in one-shot learning on \tacred and \empathe with hard and soft EM. \vspace{-3mm}}
  \label{tab:hard_soft_em}
    
\vspace{-3mm}
\end{table}

Table.~\ref{tab:hard_soft_em} shows the \ourmodel on two datasets with soft and hard EM. The temperature $\tau$ for soft EM is set as 0.5. \ourmodel with soft EM has higher performance. However, we select the hard EM as our main setting because it brings a faster training speed.


\section{Few-shot Text-style Transfer}
\label{sec:exp_text_style_transfer}
\subsection{Setting}
\label{sec:transfer_setting}
We follow the common non-parallel text style transfer setting as in ~\citep{nangi2021counterfactuals}, where each text sample $\vx$ is associated with a style label $\vy$. In the few-shot setting, the style transfer model $\pi^{s}_{\theta}: \mathcal{X} \rightarrow \mathcal{X}'$ is pre-trained on a training set, $\mathcal{D}_{train}$, which includes abundant training data (\textit{e.g.} more than 50 instances per style) for each base style $C_{b}$. After pre-training, the model parameters would be frozen with only the label embeddings updated based on a support set, $\mathcal{D}_{sup}=\{\vx,\vy\}_{i=1}^{N\times |C_{n}|}$, which includes only $N$-shot instances for each one of $|C_{n}|$ novel styles. Although \ourmodel can be easily fine-tuned on the support sets, we found the fine-tuning brings negligible performance gain in the few-shot setting. The test set $\mathcal{D}_{test}$ is sampled from both $\mathcal{D}_{train}$ and a corpus $\mathcal{D}_{train}'$ which is from the same distribution of $\mathcal{D}_{train}$. The style transfer task is to transfer text in $\mathcal{D}_{test}$ into the styles of $C_{n}$ in $\mathcal{D}_{sup}$.

\subsection{Datasets} 
\label{sec:data_style}
\empathe dataset includes around 18,000 dialogues. Each dialogue consists of a context description and an associated empathetic type. \persona includes 200,000 image captions associated with 215 personality types. Since many personalities are highly correlated in terms of their semantics, we cluster these personalities into 35 groups and manually select one type for each group. For \empathe, we use the context descriptions as the original text to be transferred and their corresponding empathetic types as the styles. For \persona, we use the image captions and their personalities. We randomly select examples of 28 empathetic types and 30 personality types in the training set and draw support sets from the rest empathetic and personalities types. We draw 0, 1 or 5 examples for each held out label. After drawing, the rest examples are considered as the test examples. For each $k$-shot, the support and test sets are drawn five times with different random seeds to avoid bias during evaluation. Our experiments would be run on all the support sets and obtain the average performance.

\subsection{Evaluation.}
We use three automatic metrics, \textit{Style-transfer Accuracy}, \textit{Self-WMD} and \textit{Perplexity}, to evaluate the accuracy of style transfer, semantic relevance and naturalness of the generated text, respectively. For \textit{Style-transfer Accuracy}, we train a BERT~\citep{devlin2018bert} classifier on styles. The averaged accuracy on target labels indicates the correctness of style transfer. \textit{Self-WMD}~\citep{kusner2015word} measures WMD between the original text and the transferred text. \textit{Perplexity} is estimated by a statistical language model in English released by~\citep{koehn2007moses}~\footnote{https://www.statmt.org/moses/RELEASE-4.0/models/cs-en/lm/}.

\subsection{Baselines.}
We compare five style transfer baselines: \textit{i)}~\rvae \citep{john2018disentangled} learns the disentangled label and content representations. \textit{ii)}~\cf \citep{nangi2021counterfactuals}, on the base of \rvae, uses a counterfactual reasoning module to control the generation of label representations. \textit{iii)}~\zf \citep{smith2019zero} is a back-translation model, which aims to deal with the zero-shot text transfer problem. The two controllable text generation baselines used in the continual few-shot setting, \textit{iv)}~\optimus and \textit{v)}~\casuallens, are extended for style transfer as well. Please refer to the Appendix~\ref{sec:implementation} and their original work for the detailed style transfer implementation.



\begin{table}[t]
    \vspace{-1mm}
\centering
  \resizebox{0.95\textwidth}{!}{%
  \begin{tabular}{|c||ccc|ccc|}
    \toprule
    \multirow{2}{*}{Methods} &
      \multicolumn{3}{c|}{\empathe} & \multicolumn{3}{c|}{\persona}\\
      & Style Accuracy$\uparrow$ & Self-WMD$\uparrow$ & Perplexity$\downarrow$ & Style Accuracy$\uparrow$ & Self-WMD$\uparrow$ & Perplexity$\downarrow$ \\
      \midrule
    \rvae & 28.49  & 90.19  & 627.49  & 27.73 &  86.06 & 796.99  \\
    \cf &  26.85  &90.82  & 657.22  & 19.69 & 87.25  & 850.86   \\
    \zf & 25.58 & 90.05  & 943.56   & 20.92 & 83.37 & 765.58  \\
    \hline
    \optimus & 27.54  &94.03 &  829.43  & 18.25 & \textbf{94.96} & 919.58 \\
    \casuallens & \textbf{39.27}  &88.74 &  1157.28  & 20.52 & 89.62 & 1059.17 \\
\hline
\ourmodel & 32.67  & \textbf{95.69}  &  \textbf{236.20}  & \textbf{49.58} & 93.44 & \textbf{311.07}  \\
                
    \bottomrule
  \end{tabular}%
  }
    \caption{ The results of one-shot style-transfer  on both datasets.      \vspace{-3mm}
}
  \label{tab:style_transfer}
    \vspace{-3mm}
\end{table}

\subsection{Inference.} 
The inference of \ourmodel for the text style transfer differs from the inference for continual few-shot learning. Given a new style, we start with sampling a name representation and text representations from the posterior label distribution, $\qP(\rmZ_y | X, y)$, conditioned on its associated name phrase and text sequences in the support set, respectively. Then, we create the label representation $\vz_y$ for the new style by averaging its associated text representations and its name representation. The content representations are sampled from the posterior content distribution, $\qP(\rmZ_c | X, C)$, conditioned on the text to be style-transferred. After feeding the content representations and the representations of target styles to the generator, we obtain the most likely outputs by beam-search.

\subsection{Main Results and Discussions.}

The results in Table~\ref{tab:style_transfer} show that our method performs better than all baselines in terms of all metrics except \textit{Style Accuracy} on \empathe and \textit{Self-WMD} on \persona. An ideal style transfer model should find a good balance in terms of all three evaluation metrics. Though \casuallens and \optimus can achieve the best on a single metric, they fail to perform well across all the metrics. We inspect that \casuallens performs poorly on preserving the content of the original sentence while \optimus performs poor on style transfer and basically replicates the original sentences in \persona dataset. In contrast, the average ranking of \ourmodel on three metrics are highest among all baselines. Our model performs particularly well in terms of semantic relevance and naturalness while still keeping high accuracies of style transfer. Other methods that utilize disentanglement learning, including \rvae, \cf and \casuallens, often perform well on one metric while lose on the other metrics. We conjecture this is due to their methods do not fully disentangle the representations so they can not balance well between content preservation and style transfer.


\subsection{Complete Automatic Evaluation Results of Style Transfer on two Datasets}
The full results of automatic evaluation on Empathetic dataset and Personality dataset are presented in Table \ref{tab:style_empath} and Tabel \ref{tab:style_persona} respectively. Overall, on all few-shot settings, our method perform the best in terms of the average rank among all baselines. Although on \empathe dataset, \rvae and \casuallens outperform our method in term of the Style Accuracy. Through inspection, we found that \rvae and \casuallens tend to overfit to support set after finetuning merely on a small number of training instances. For example, \rvae tends to copy the text from support set, which gains higher Style Transfer Accuracy. But this effect makes the Perplexity and Self-WMD of \rvae and \casuallens decreasing from zero-shot to five-shot learning. In contrast, \ourmodel performs steady across different(zero/few-shot) settings. The Style Accuracy of \ourmodel is increasing without losing performance on content preservation and naturalness of sentences.
\begin{table*}[ht]
    \vspace{-1mm}
\centering
  \resizebox{0.95\textwidth}{!}{%
  \begin{tabular}{|c|ccc|ccc|ccc|ccc|}
    \toprule
    \multirow{2}{*}{Methods} &
      \multicolumn{3}{c|}{Style Accuracy$\uparrow$} &
      \multicolumn{3}{c|}{Self-WMD$\uparrow$}& \multicolumn{3}{c|}{Perplexity$\downarrow$} & \multicolumn{3}{c|}{AVG Rank$\downarrow$}\\
      & 0  & 1 & 5  & 0 & 1  & 5 & 0 & 1  & 5   & 0 & 1  & 5  \\
      \midrule
    \rvae & 33.99 &28.49 & 41.90 & 91.39  & 90.19& 90.03 & 796.47  & 627.49  & 896.19   & 3.3 &	3.00	& 3.33  \\
    \cf & 20.39 & 26.85 & 26.99  &93.95 & 90.82 & 90.75  & 825.80  & 657.22  & 858.61 & 4.67	& 3.67	& 4 \\
    \zf & 21.85 & 25.58 & 26.60  &93.64  & 90.05 & 89.51 & 785.29  & 943.56  & 609.68  & 3.67 &	5.33 &	4.33 \\
    \optimus & 26.26 & 27.54 & 27.50  &94.24  & 94.03 & 93.91 & 814.45  & 829.43  & 826.63  & 3.33 &	3.33 &	3 \\
        \casuallens & 34.53 & 39.27  & 41.32  &89.74 & 88.74 & 88.42  & 1236.31 & 1157.28 & 1332.15   & 4.33	&4.33	&4.67  \\
\hline
\ourmodel & 32.84 & 32.67 & 34.55 &95.68  & 95.69 & 95.70 & 233.03  & 236.20  & 233.66 & 1.67 &	1.33	& 1.67\\
    \bottomrule
  \end{tabular}%
  }
    \caption{The results of zero, one and five-shot learning of style transfer on \empathe dataset.
     \vspace{-3mm} }
  \label{tab:style_empath}
    \vspace{-3mm}
\end{table*}

\begin{table*}[ht]
    \vspace{-1mm}
\centering
  \resizebox{0.95\textwidth}{!}{%
  \begin{tabular}{|c|ccc|ccc|ccc|ccc|}
    \toprule
    \multirow{2}{*}{Methods} &
      \multicolumn{3}{c|}{Style Accuracy$\uparrow$} &
      \multicolumn{3}{c|}{Self-WMD$\uparrow$}& \multicolumn{3}{c|}{Perplexity$\downarrow$} & \multicolumn{3}{c|}{AVG Rank$\downarrow$}\\
      & 0  & 1 & 5  & 0 & 1  & 5 & 0 & 1  & 5 & 0 & 1  & 5  \\
      \midrule
    \rvae & 21.84 &27.73 & 21.20 & 91.94  & 86.06& 86.91 & 832.40  & 796.99  & 1202.90 & 3& 3.33 &	4.33  \\
    \cf & 19.88 & 19.69 & 18.72 &95.62 & 87.25 & 87.92  & 852.62  & 850.86  & 1205.90& 2.67 &	4.33 & 5\\
    \zf & 16.98 & 20.92 & 21.80 &87.45  & 83.37 & 83.63 & 643.00  & 765.58  & 891.42& 4.67 & 3.67 &	3.67 \\
    \optimus & 17.79 & 18.25 & 19.21 &95.32  & 94.96 & 95.00 & 966.29  & 919.58  & 955.78 & 4 &	4 &	3\\
    \casuallens & 19.46& 20.52 & 22.69 &91.30  & 89.62 & 89.15 & 1238.83  & 1059.17  & 1567.96 & 5.00 &	4.33 & 3.67\\
\hline
\ourmodel & 50.64 & 49.58 & 55.51 &93.69  & 93.44 & 93.46 & 322.03  &311.07  & 304.36 & 1.67 & 1.33 & 1.33\\
                
    \bottomrule
  \end{tabular}%
  }
    \caption{The results of zero, one and five-shot learning of style transfer on \persona dataset.
     \vspace{-3mm} }
  \label{tab:style_persona}
    \vspace{-3mm}
\end{table*}

\subsection{Human evaluation result}
We hire three crowd-workers to rate the sentences with score from 1-5 to indicate whether the the generated sentences belong to the target styles and whether the content of generated sentences are consistent with the original sentences. To evaluate naturalness, we follow the evaluation setting in~\cite{mir2019evaluating} to let the crowd-workers distinguish the human generated sentences from the model generated sentences. The naturalness score in Tab. \ref{tab:human} indicates successful rate of distinguishing the sentences. The easier the sentence is distinguished, the less natural the sentence is. We achieve far superior performance in terms of both Content Preservation and Style Transfer metrics. Although on Naturalness, our method only ranks third. We conjecture that the generated sentences by \ourmodel is usually longer than the original sentences. The crowd-workers could easily grasp this pattern and distinguish the sentences. Besides, with Naturalness metric, the gap between different methods are actually insignificant, which are all close to 50\%.
\begin{table}
\begin{tabular}{|c|cccc|} \toprule
{Method} & {Content$\uparrow$} &{Style$\uparrow$} &{Nature$\downarrow$} & {Rank$\downarrow$}\\ \midrule
\cf & 1.33 & 2.26  & 0.45 & 2.67 \\ 
\rvae  & 1.20 & 2.15 & \textbf{0.44} & 3.33 \\ 
\zf & 1.17 & 2.01 & 0.57 & 5.33\\ 
\optimus  & 2.33 & 1.85 & 0.56 & 4 \\ 
\casuallens  & 2.04 & 2.24 & 0.58 & 4 \\ 
\textbf{\ourmodel}  & \textbf{2.44} & \textbf{2.40} & 0.52 & \textbf{1.67} \\ 
    \bottomrule
\end{tabular}
\caption{Human evaluation results, evaluated by content preservation (Content), style transfer correctness (Style), naturalness (Nature), and average rank of the three criterions (Rank).}
\label{tab:human}
\end{table}

\subsection{Generated Examples of Style Transfer}
\begin{table}[ht]
\centering
 \resizebox{0.9\textwidth}{!}{%
\begin{tabular} {|c|c|} \toprule
Methods & Original Style: prepared $\rightarrow$ Target Style: anticipating \\ \midrule
\zf &  \parbox[c]{13cm}{i have m so scared of spiders. i can't stand those things!} \\ \midrule
\cf &  \parbox[c]{13cm}{i was schocked to see my favorite band wasnt coming to my city this tour} \\ \midrule
\rvae & \parbox[c]{13cm}{i cannot wait until next month. i had a feeling my birthday was.}\\ \midrule
\optimus & \parbox[c]{13cm}{I thought I didn't planned for my job interview at jobster trip . I felt like going off}\\ \midrule
\casuallens & \parbox[c]{13cm}{I felt very apprehensive when I went to my interview} \\ \midrule
\textbf{\ourmodel} & \parbox[c]{13cm}{I felt really good at my job interview at work today I felt I did well at the job I worked out for when I saw I had done well in my interview for the position I was looking forward to doing at that time .} \\ \bottomrule
\end{tabular}
\caption{The style transfer results of different models trained on dataset \empathe with one-shot learning setting. The original sentence, "i felt like i did well at my job interview yesterday. i went in feeling confident", is transferred from the original style "prepared" to the target style "anticipating".}

\label{tab:empathe_examples}
}

\end{table}

\begin{table}
 \resizebox{0.9\textwidth}{!}{%
\begin{tabular} {|c|c|} \toprule
Methods & Original Style: appreciative $\rightarrow$ Target Style: angry \\ \midrule
\cf &  \parbox[c]{13cm}{the amount of shadows in the middle of.} \\ \midrule
\rvae & \parbox[c]{13cm}{the amount of players in the left of building.}\\ \midrule
\zf &  \parbox[c]{13cm}{these mountains just look so nice! i would love to see them.} \\ \midrule
\optimus & \parbox[c]{13cm}{fl in the fissip , the fissile columns in theTyphris ' windows in the Sky .}\\ \midrule
\casuallens & \parbox[c]{13cm}{the horizon is filled with dazzling colors .}\\ \midrule
\textbf{\ourmodel} & \parbox[c]{13cm}{Look at the splashes on the rocks in the middle of the street , I hate looking at rocks . They're so ugly looking , and I can't stand to look at them anymore .} \\ \bottomrule
\end{tabular}
\caption{The style transfer results of different models trained on dataset \persona with one-shot learning setting. The original sentence, "Look at the fissures in the strata columns, beautiful.", is transferred from the original style "appreciative" to the target style "angry".}

\label{tab:persona_examples}
}

\end{table}
Table \ref{tab:empathe_examples} and \ref{tab:persona_examples} depict the examples of generate examples of different style transfer methods trained on \empathe and \persona, respectively.

\section{Related Work Supplementary}

\noindent\textbf{Data Augmentation.}
Data augmentation (DA) encompasses methods of increasing training data diversity without directly collecting more data~\cite{abs-2105-03075dasurvey}, which can be roughly categorized into (1) rule-based methods~\cite{wei2019eda}, (2) example interpolation methods~\cite{ZhangCDL18mixup}, and (3) model-based methods~\cite{wu2019conditional,sennrich2015improving,anaby2020not,kumar2020data,lee2021neural,shiri2022paraphrasing}. Data augmentation generally encourages better performance in low-resource scenarios, such as few-shot learning~\cite{kumar-etal-2019-closer} and low-resource language learning~\cite{xia-etal-2019-generalized}. Although data augmentation has been well applied in many tasks~\cite{abs-2105-03075dasurvey}, there has been limited work on DA for conditional text generation~\cite{feng-etal-2020-genaug}. 

\noindent\textbf{Continual Few-shot Learning.}
The primary challenge addressed in continual learning literature is overcoming catastrophic forgetting~\cite{french1999catastrophic,biesialska-etal-2020-continual,wu2022pretrained}, Various approaches have been proposed to tackle the forgetting problem, e.g., rehearsal-based methods~\cite{han2020continual,dAutumeRKY19,li2021total,abs-2101-01926}, regularization-based methods~\cite{li-etal-2019-continuous,huang-etal-2021-continual}, and dynamic architecture methods~\cite{ke-etal-2021-adapting,LinMF20}. Continual few-shot learning is an even more challenging yet realistic setting which encourages learners the quick adaptation ability during learning~\cite{brahma2021hypernetworks,pmlr-v119-yoon20b}. Comparing to the numerous researches out of NLP applications~\cite{pmlr-v139-yap21a,pmlr-v119-yoon20b,DongHTCWG21}, continual few-shot language learning is still an under-explored area~\cite{brahma2021hypernetworks}. 


\section{Proofs}

\subsection{Discussion about $\epsilon$-Disentangled}
\label{sec:disentangled}
To achieve information purity, the learned models should follow the structure illustrated in Fig.\ref{fig:DAG}(a) that there is no dependency between $C$ and $\rmZ_y$, and similarly no dependency between $y$ and $\rmZ_c$. However, prior works on disentangled representation learning regularize the models by minimizing mutual information $I(\rmZ_c,\rmZ_y)$ between $\rmZ_c$ and $\rmZ_y$ such that $\rmZ_c \independent \rmZ_y$ when $I(\rmZ_c,\rmZ_y)=0$~\cite{cheng2020VIDisentangled,wang2021desiderata}. In another word, prior works only require that there is no edge between $\rmZ_c$ and $\rmZ_y$ in the Bayesian model. However, this does not imply $I(\rmZ_c,y)=0$ and $I(\rmZ_y,C)=0$. In the trained models, $C$ can still be the shared parent or child of two independent random variables using the regularization from the prior works. In addition, the independence assumption between $\rmZ_c$ and $\rmZ_y$ does not always hold in practice. For example, if $\rmZ_y$ is a random variable for emotion categories and $\rmZ_c$ represents events influencing emotions, they are causally dependent. Forcing the independent assumption may deteriorate model performance.

To address this limitation, we propose to regularize the priors of latent variables for encouraging information purity. If we have a close look at $I(\rmZ_y, y) = \int p(\rmZ_y, y) \log \frac{p(\rmZ_y, y)}{p(\rmZ_y) p(y)}$, which is simplified to $ \int p(y | \rmZ_y) p(\rmZ_y) \log \frac{p(\rmZ_y | y)}{p(\rmZ_y)}$, a high mutual information expects $p(\rmZ_y) > 0$ whenever $p(\rmZ_y | y)$ is high. Similarly, if we aim for an extremely small $I(\rmZ_y, C)$, we expect a low $p(\rmZ_y)$ or $p(\rmZ_y) = 0$ whenever $p(\rmZ_y | C) > p(\rmZ_y)$. If we design the priors in the way that their dense regions are not overlapped, we achieve information purity by maximizing the corresponding mutual information. 

We do not require \textbf{absolute continuity} for the associated divergence measure because when the priors are $\epsilon$-disentangled with 
a fairly low $\epsilon$, one of the priors would have zero probability in the regions where the other prior has positive supports.

\subsection{Latent Variable Non-identifiability}
\label{sec:non-identifiability}
\newcite{wang2020posterior} introduce the concept of latent variable non-identifiability and shows that it leads to posterior collapse.
\begin{definition}[Latent variable non-identifiability \cite{wang2020posterior}]
\label{def:non_identifiable}
Given a likelihood function $\pT(\rmX| \rmZ ;\vtheta)$ with parameters $\vtheta = \hat{\vtheta}$ and a dataset $\gD = \{\vx_1, ...,\vx_n\}$, the latent variables $\rmZ$ are non-identifiable if $p(\gD | \rmZ = \vz ; \hat{\vtheta}) = p(\gD | \rmZ = \vz' ; \hat{\vtheta})$ $\forall \vz, \vz' \in \gZ$, where $\gZ$ denotes the space of latent variable values. As a consequence, $p(\gD | \rmZ; \hat{\vtheta}) = p(\gD ; \hat{\vtheta})$.
\end{definition}

For the cases with more than one random variables (vectors), we extend this idea for latent condtional non-identifiability.
\begin{definition}[Latent variable conditional non-identifiability]
\label{def:conditional_non_identifiable}
Given a likelihood function $\pT(\rmX| \rmZ_a, \rmZ_b ;\vtheta)$ with parameters $\vtheta = \hat{\vtheta}$ and a dataset $\gD = \{\vx_1, ...,\vx_n\}$, the latent variables $\rmZ_a$ are non-identifiable conditioned on $\rmZ_b$ if $p(\gD | \rmZ_a, \rmZ_b ; \hat{\vtheta}) = p(\gD | \rmZ_b ; \hat{\vtheta})$. 
\end{definition}

\begin{proposition}
\label{pro:conditional_non_identifiable}
Given a likelihood function $\pT(\rmX| \rmZ_a, \rmZ_b ;\vtheta)$ with parameters $\vtheta = \hat{\vtheta}$ and a dataset $\gD = \{\vx_1, ...,\vx_n\}$, the latent variables $\rmZ_a$ are non-identifiable conditioned on $\rmZ_b$ if $p(\rmZ_a| \rmZ_b ; \hat{\vtheta}) = 1$.
\end{proposition}

\textit{Proof}:

Given $p(\rmZ_a| \rmZ_b ; \hat{\vtheta}) = 1$,
\begin{align}\nonumber
\begin{split}
p(\gD | \rmZ_b ; \hat{\vtheta}) p(\rmZ_a| \rmZ_b ; \hat{\vtheta}) = p(\gD | \rmZ_a, \rmZ_b ; \hat{\vtheta}) \\
p(\gD | \rmZ_b ; \hat{\vtheta}) = p(\gD | \rmZ_a, \rmZ_b ; \hat{\vtheta})
\end{split}
\end{align}

For all $\vx_i$ in $\gD$, if $p(\vx_i |\rmZ_a = \vz, \rmZ_b = \vz ) > p(\vx_i |\rmZ_a = \vz, \rmZ_b = \vz' )$ with $\vz \neq \vz'$ for all $\vz, \vz' \in \gZ$, where $\gZ$ is the space of latent variable values, $p(\rmZ_a | \rmZ_b)$ is high because both random variable vectors are almost a copy to each other. In this case, if $p_a(\rmZ_a)$ and $p_b(\rmZ_b)$ share the same dense regions or even the same, such a conditional non-identifiable case will not be penalized during training. In contrast, if $p_a(\rmZ_a)$ and $ p_b(\rmZ_b)$ are $\epsilon$-disentangled with a small $\epsilon$, the parameters leading to the conditional non-identifiable cases are disencouraged by receiving zero or a low likelihood $p(\vx_i |\rmZ_a = \vz, \rmZ_b = \vz ; \vtheta) p_a(\vz) p_b(\vz)$.

\subsection{Proofs for VAE with Disentanglement Priors}
\label{sec:ELBO}
The main difficulty of maximum likelihood learning for the optimization problem \eqref{eq:opt_problem} is that the marginal probability of data $p(\rmX|C,y)$
under the model is intractable. We apply the variational techniques to derive the ELBO for the optimization problem \eqref{eq:opt_problem}, whose constraint is removed by introducing the disentanglement priors. 

In the VAE framework, we adopt variational distributions to approximate true distributions~\cite{kingma2019vaeBook}, which ends up maximizing an ELBO. More specifically, we introduce a variational posterior $q_{\phi}(\rmZ_c, \rmZ_y | \rmX, C, y)$ to approximate the true posterior $\pT(\rmZ_c, \rmZ_y | \rmX, C, y)$, and derive the ELBO for $\pT(\rmX|C,y)$ in Sec. ~\ref{appendix:elbo}:
\begin{align} \nonumber
    &\mathbb{E}_{q_{\vPhi}(\rmZ_c, \rmZ_y | X, C, y)}[\log p_{\vTheta}(X, \rmZ_c, \rmZ_y |C, y ) \\
    &- \log q_{\vPhi}(\rmZ_c, \rmZ_y | X, C, y)]
\end{align}
 
We show in Sec.~\ref{appendix:elbo_decomposition}
that the ELBO objective is further decomposed into:
\begin{small}
\begin{align}
\label{eq:elbo_decompose}
    \begin{split}
        &\overbrace{\mathbb{E}_{q_{\vPhi}(\rmZ_c, \rmZ_y | \rmX, C, y)}[\log p_{\vTheta}(\rmX| \rmZ_c, \rmZ_y )]}^{\gL_r} \\
        &- \KL(\qP(\rmZ_c | \rmX, C)  \| \pT(\rmZ_c | C))  \\ 
        &- \KL(\qP(\rmZ_y | \rmX, y)  \| \pT(\rmZ_y | y))
    \end{split}
\end{align}
\end{small}
where the first term is referred to as the reconstruction loss $\gL_r$, the other terms constitute regularizers. 


\label{appendix:proofs}
\subsubsection{Evidence lower bound (ELBO)}
\label{appendix:elbo}
\begin{small}
\begin{displaymath}
    \log p(\rmX) \geq  \mathbb{E}_{q_{\vPhi}(\rmZ_c, \rmZ_y | \rmX, C, y)}[\log p_{\vTheta}(\rmX, \rmZ_c, \rmZ_y |C, y ) - \log q_{\vPhi}(\rmZ_c, \rmZ_y | \rmX, C, y)]
\end{displaymath}
\noindent
\\
\textit{Proof:}
\begin{align}\nonumber
\begin{split}
    &\KL (q_{\vPhi} (\rmZ_c, \rmZ_y | \rmX, C,y) \| p_{\vTheta} (\rmZ_c, \rmZ_y | \rmX, C,y))\\
    =& - \int \qP (\rmZ_c, \rmZ_y | \rmX, \cons) \Big[\log \frac{ p_{\vTheta}(\rmZ_c, \rmZ_y |\rmX,C, y ))}{\qP(\rmZ_c, \rmZ_y | \rmX, \cons)}\Big] d\rmZ_c d\rmZ_y\\
    =& - \int \qP (\rmZ_c, \rmZ_y | \rmX, \cons) \Big[\log \frac{ p_{\vTheta}(\rmX, \rmZ_c, \rmZ_y |C, y ))}{\qP(\rmZ_c, \rmZ_y | \rmX, \cons)\pT (\rmX | \cons)}\Big] d\rmZ_c d\rmZ_y\\
    =& - \int \qP (\rmZ_c, \rmZ_y | \rmX, \cons) \Big[\log \frac{ p_{\vTheta}(\rmX, \rmZ_c, \rmZ_y |C, y ))}{\qP(\rmZ_c, \rmZ_y | \rmX, \cons) } - \log \pT (\rmX | \cons)\Big] d\rmZ_c d\rmZ_y
\end{split}
\end{align}
$\KL (q_{\vPhi} (\rmZ_c, \rmZ_y | \rmX, C,y) \| p_{\vTheta} (\rmZ_c, \rmZ_y | \rmX, C,y)) \geq 0$, therefore:
\begin{align}\nonumber
\begin{split}
\log \pT (\rmX | \cons) \geq& \int \qP (\rmZ_c, \rmZ_y | \rmX, \cons) \Big[\log \frac{p_{\vTheta}(\rmX, \rmZ_c, \rmZ_y |C, y ))}{\qP(\rmZ_c, \rmZ_y | \rmX, \cons) }\Big] d\rmZ_c d\rmZ_y\\
=& \mathbb{E}_{q_{\vPhi}(\rmZ_c, \rmZ_y | \rmX, C, y)}[\log p_{\vTheta}(\rmX, \rmZ_c, \rmZ_y |C, y ) - \log q_{\vPhi}(\rmZ_c, \rmZ_y | \rmX, C, y)]
\end{split}
\end{align}
\end{small}
\subsubsection{ELBO decomposition}
\label{appendix:elbo_decomposition}
\begin{small}
If $\rmZ_c \independent \rmZ_y | C$, then
\begin{align}\nonumber
    \text{ELBO} =& \mathbb{E}_{q_{\vPhi}(\rmZ_c, \rmZ_y | \rmX, C, y)}[\log p_{\vTheta}(\rmX| \rmZ_c, \rmZ_y )] - \KL(\qP(\rmZ_c | \rmX, C)  \| \pT(\rmZ_c | C)) - \KL(\qP(\rmZ_y | \rmX, y)  \| \pT(\rmZ_y | y)).
\end{align}
\\
\noindent
\textit{Proof:}

Let $p_{\vTheta}(\rmX,\rmZ_c, \rmZ_y| C,y)=\pT(\rmX |\rmZ_{c},\rmZ_y)\pT(\rmZ_{c}|C)\pT(\rmZ_y|y)$, then
\begin{align}\nonumber
    \begin{split}
&\mathbb{E}_{q_{\vPhi}(\rmZ_c, \rmZ_y | \rmX, C, y)}[\log p_{\vTheta}(\rmX, \rmZ_c, \rmZ_y |C, y ) - \log q_{\vPhi}(\rmZ_c, \rmZ_y | \rmX, C, y)]\\
=& \int \qP (\rmZ_c, \rmZ_y | \rmX, \cons) \Big[\log \frac{\pT(\rmX |\rmZ_{c},\rmZ_y)\pT(\rmZ_{c}|C)\pT(\rmZ_y|y))}{\qP(\rmZ_c, \rmZ_y | \rmX, \cons) }\Big] d\rmZ_c d\rmZ_y\\
=&\mathbb{E}_{q_{\vPhi}(\rmZ_c, \rmZ_y | \rmX, C, y)}[\log p_{\vTheta}(\rmX| \rmZ_c, \rmZ_y )] + \int \qP (\rmZ_c, \rmZ_y | \rmX, \cons) \Big[\log \frac{\pT(\rmZ_{c}|C)\pT(\rmZ_y|y))}{\qP(\rmZ_c, \rmZ_y | \rmX, \cons) }\Big] d\rmZ_c d\rmZ_y\\
=& \mathbb{E}_{q_{\vPhi}(\rmZ_c, \rmZ_y | \rmX, C, y)}[\log p_{\vTheta}(\rmX| \rmZ_c, \rmZ_y )] + \int \qP(\rmZ_c, \rmZ_y  | \rmX, C, y) \Big[\log \frac{\pT(\rmZ_{c}|C)\pT(\rmZ_y|y))}{\qP(\rmZ_c |\rmZ_y, \rmX, C) \qP(\rmZ_y | \rmX, y)} \Big] d\rmZ_c d\rmZ_y \\
    \end{split}
\end{align}
Because $\rmZ_c \independent \rmZ_y | C$, $q_{\vPhi}(\rmZ_c, \rmZ_y | \rmX, C, y) = \qP(\rmZ_c | \rmX, C) \qP(\rmZ_y | \rmX, y)$, thus
\begin{align}\nonumber
    \begin{split}
& \mathbb{E}_{q_{\vPhi}(\rmZ_c, \rmZ_y | \rmX, C, y)}[\log p_{\vTheta}(\rmX| \rmZ_c, \rmZ_y )] + \int \qP(\rmZ_c | \rmX, C) \qP(\rmZ_y | \rmX, y) \Big[\log \frac{\pT(\rmZ_{c}|C)\pT(\rmZ_y|y))}{\qP(\rmZ_c | \rmX, C) \qP(\rmZ_y | \rmX, y)} \Big] d\rmZ_c d\rmZ_y\\
=& \mathbb{E}_{q_{\vPhi}(\rmZ_c, \rmZ_y | \rmX, C, y)}[\log p_{\vTheta}(\rmX| \rmZ_c, \rmZ_y )]-\KL(\qP(\rmZ_c | \rmX, C) \qP(\rmZ_y | \rmX, y)  \| \pT(\rmZ_c | C)\pT(\rmZ_y | y))\\
=& \mathbb{E}_{q_{\vPhi}(\rmZ_c, \rmZ_y | \rmX, C, y)}[\log p_{\vTheta}(\rmX| \rmZ_c, \rmZ_y )] - \KL(\qP(\rmZ_c | \rmX, C)  \| \pT(\rmZ_c | C)) - \KL(\qP(\rmZ_y | \rmX, y)  \| \pT(\rmZ_y | y)).
    \end{split}
\end{align}
\end{small} \\
Note that, the last step is derived by applying the chain rule of KL divergence.

\subsubsection{Derivation of the regularization term for latent label representations}
\label{appendix:label_representation}
If $\qP(\rmZ_y |\rmX, y) = \mathcal{N}(\rmZ_y ; \vMu^q_y, \text{diag}(\vSigma^2)_y)$ and $\pT(\rmZ_y | y) = \mathcal{N}(\rmZ_y ; \vMu^p_y, \lambda_y\rmI)$,
where $\vMu^q_y, \log \vSigma_y = \text{LabelEncoder}(\rmX)$ and $\vMu_y^p = \rmW_y \Phi(l)$, then we have:
\begin{displaymath}
    \KL(\qP(\rmZ_y | \rmX, y)  \| \pT(\rmZ_y | y)) = \frac{1}{2\lambda_y}\|\rmZ_y - \vMu_y^p\|^2 - \log \vSigma_y^q + \text{const}
\end{displaymath}

\noindent
\textit{Proof:}
Let $\rmZ_y = \vMu^q_y + \bm{\sigma}_y \odot \bm{\epsilon}_y$, where $\bm{\epsilon}_y$ is drawn from $\mathcal{N}(0,\rmI)$.

\begin{align}\nonumber
\begin{split}
    \KL(\qP(\rmZ_y | \rmX, y)  \| \pT(\rmZ_y | y)) =& \mathbb{E}_{\qP(\rmZ_y | \rmX, y)}\big[\log \qP(\rmZ_y | \rmX, y)  - \log \pT(\rmZ_y | y)\big]\\
    =& \mathbb{E}_{p(\bm{\epsilon}_y)}\big[\log \qP(\rmZ_y | \rmX, y)  - \log \pT(\rmZ_y | y)\big]
\end{split}
\end{align}

$\pT(\rmZ_y | y) = \mathcal{N}(\rmZ_y ; \vMu^p_y, \lambda_y\rmI)$, thus
\begin{align}\nonumber
\begin{split}
\log \pT(\rmZ_y | y) = -\frac{1}{2\lambda_y}\|\rmZ_y - \vMu_y^p\|^2 + \text{const}
\end{split}
\end{align}

Using the reparameterization trick,
\begin{align}\nonumber
\begin{split}
\log \qP(\rmZ_y | \rmX, y) =& \log p(\bm{\epsilon}_y) - \log |\text{det}\big(\frac{\partial \rmZ_y}{\partial \bm{\epsilon}_y}  \big) |\\
= &\log \mathcal{N}(\bm{\epsilon}_y ; 0, \rmI) - \log\vSigma^q_y
\end{split}
\end{align}

Put them together
\begin{displaymath}
    \mathbb{E}_{p(\bm{\epsilon}_y)}\big[\log \qP(\rmZ_y | \rmX, y)  - \log \pT(\rmZ_y | y)\big] = \frac{1}{2\lambda_y}\|\rmZ_y - \vMu_y^p\|^2 - \log \vSigma_y^q + \text{const}
\end{displaymath}

\subsubsection{Derivation of the regularization term for latent content representations}
We assume
$
    \qP(\rmZ_c |\rmX, C) = \mathcal{N}(\rmZ_c ; \vMu^q_{c}, \text{diag}(\vSigma_c^2))  $ and $
    \pT(\rmZ_c | C) = \sum_{k=1}^K p(M = k) \mathcal{N}(\rmZ_c ; \vMu^p_{c,k}, \lambda_c\rmI)\\
$ then we have
\begin{align}\nonumber
\begin{split}
         \KL(\qP(\rmZ_c | \rmX, C)  \| \pT(\rmZ_c | C)) = &\sum_{k=1}^K p(M = k|\rmZ_c)\big[\frac{1}{2\lambda_c}\|\rmZ_c - \vMu_{c,k}^p\|^2 \big] - \log \vSigma_{c} + \text{const}
\end{split}
\end{align}

 \noindent
\textit{Proof:}

Let $\rmZ_c = \vMu^q_c + \bm{\sigma}_c \odot \bm{\epsilon}_c$, where $\bm{\epsilon}_c$ is drawn from $\mathcal{N}(0,\rmI)$.
\begin{align}\nonumber
\begin{split}
\KL(\qP(\rmZ_c | \rmX, C)  \| \pT(\rmZ_c | C)) = & \mathbb{E}_{\qP(\rmZ_c | \rmX, C)}\big[\log \qP(\rmZ_c | \rmX, C) -  \log \pT(\rmZ_c | C)\big]\\
 = &  \mathbb{E}_{p(\bm{\epsilon}_c)}\big[\log \qP(\rmZ_c | \rmX, C) -  \log \pT(\rmZ_c | C)\big]\\
 = &  \mathbb{E}_{p(\bm{\epsilon}_c)}\big[\log \qP(\rmZ_c | \rmX, C)\big] -  \log \pT(\rmZ_c | C)
\end{split}
\end{align}

Using the reparameterization trick,
\begin{align}\nonumber
\begin{split}
\mathbb{E}_{p(\bm{\epsilon}_c)}\big[\log \qP(\rmZ_c | \rmX, y)\big] = - \log\vSigma^q_c
\end{split}
\end{align}

It remains to estimate $\log \pT(\rmZ_c | C)$, which is a Gaussian mixture. Let $\gamma_k \in \{0,1\}$ indicate the $k$th component of $\vz$, the likelihood function for $\vz$ takes the form 
\begin{displaymath}
    \pT (\vz, \bm{\gamma}) = \prod_{k=1}^K p(M=k)^{\gamma_k}\mathcal{N}(\vz | \vMu^p_{c,k}, \lambda_c\rmI)^{\gamma_k}
\end{displaymath}

In the work, we consider using EM~\cite{bishop2006pattern}, which estimates the expected value of the complete log likelihood function given by
\begin{align}\nonumber
\label{eq:complete_likelihood}
    \mathbb{E}_{\bm{\gamma}} \big[ \log \pT (\vz, \bm{\gamma}) \big] =& \sum_{k=1}^K \mathbb{E}(\gamma_k)\{\log p(M=k) + \log \mathcal{N}(\vz | \vMu^p_{c,k}, \lambda_c\rmI)\}\\
    =& \sum_{k=1}^K \mathbb{E}(\gamma_k)\{-\frac{1}{2\lambda_c}\|\rmZ_c - \vMu_{c,k}^p\|^2\}\} + \text{const}
\end{align}
where $\mathbb{E}(\gamma_k) = p(M = k | \vz_c) = \frac{p(M = k)\mathcal{N}(\vz | \vMu^p_{c,k}, \lambda_c\rmI)}{\sum_{j=1}^K p(M = j)\mathcal{N}(\vz | \vMu^p_{c,j}, \lambda_c\rmI)}$, estimated in the E-step.

For hard EM:

\textbf{E-step}. For each latent content representation $\vz_c$, the most likely component Gaussian is given by $k^* = \arg\max_k \pT(M = k) \mathcal{N}(\rmZ_c ; \vMu^p_{c,k}, \lambda_c\rmI)$.

\textbf{M-step}. Put the estimated $k^*$ into Eq. \eqref{eq:complete_likelihood}, this step aims to optimize
\begin{align} \nonumber
    \sum_{k=1}^K \gamma_{k^*}\{-\frac{1}{2\lambda_c}\|\rmZ_c - \vMu_{c,k}^p\|^2\} + \text{const}
\end{align}

Put them together, we have 
\begin{align}\nonumber
\begin{split}
         \mathbb{E}_{p(\bm{\epsilon}_c)}\big[\log \qP(\rmZ_c | \rmX, C)\big] -  \log \pT(\rmZ_c | C) = &\sum_{k=1}^K p(M = k|\rmZ_c)\big[\frac{1}{2\lambda_c}\|\rmZ_c - \vMu_{c,k}^p\|^2 \big] - \log \vSigma_{c} + \text{const}
\end{split}
\end{align}



\section{Model Regularization}
\label{appendix:regularization}
\subsection{HSIC Regularization}
\label{appendix:hsic}
In each batch, the model collects the latent representations $(\rmZ_c, \rmZ_y)$, which are a content representation matrix and a label representation matrix respectively. We apply the linear kernel to build a Gram matrix $\rmK_c = \rmZ_c \rmZ^{\intercal}_c$ for content and a Gram matrix $\rmK_y = \rmZ_y \rmZ^{\intercal}_y$ for labels. The HSIC metric is computed as
\begin{equation}
    \text{HSIC}(\rmZ_c, \rmZ_y) = \frac{1}{m^2} \text{trace}(\rmK_c \rmH \rmK_y \rmH)
\end{equation}
where $\rmH = \rmI - \frac{1}{m}\mathbf{1}\mathbf{1}^\intercal$ and $m$ is the size of the batch. Alternatively, we can try the Gaussian Kernel for both types of representations.
\subsection{MMD Regularization}
The MMD divergence is given by~\cite{gretton2012mmd}: 
\begin{multline}
    \text{MMD}(Z_c, Z_y) = \frac{1}{m^2}\sum_{i=0}^m \sum_{j=0}^m k(\vz_i^c, \vz_j^c)
    - \frac{2}{m^2}\sum_{i=0}^m \sum_{j=0}^m k(\vz_i^c, \vz_j^y) +  \frac{1}{m^2}\sum_{i=0}^m \sum_{j=0}^m k(\vz_i^y, \vz_j^y)
\end{multline}
where $k(\cdot,\cdot)$ is a kernel function, whereby we choose the linear kernel in our experiments. Maximizing MMD increases the similarity of the latent representations of the same type, while decreases the similarity of the latent representations across types.


\end{document}